\documentclass[lettersize,journal]{IEEEtran}
\usepackage{amsmath,amsfonts}
\usepackage{algorithmic}
\usepackage{algorithm}
\usepackage{array}
\usepackage[caption=false,font=footnotesize,labelfont=rm,textfont=rm]{subfig}
\usepackage{textcomp}
\usepackage{stfloats}
\usepackage{url}
\usepackage{verbatim}
\usepackage[T1]{fontenc}
\usepackage{graphicx}
\usepackage{cite}
\usepackage{booktabs}
\usepackage{multirow}
\usepackage{color}
\usepackage[colorlinks=true,linkcolor=blue,citecolor=blue,urlcolor=magenta]{hyperref}

\makeatletter 
\let\myorg@bibitem\bibitem
\def\bibitem#1#2\par{%
\@ifundefined{bibitem@#1}{%
\myorg@bibitem{#1}#2\par
}{%
\begingroup
\color{\csname bibitem@#1\endcsname}%
\myorg@bibitem{#1}#2\par
\endgroup
}%
}
\makeatother 

\begin{document}

\title{\huge{Collaborative Learning of Scattering and Deep Features for SAR Target Recognition with Noisy Labels}}

\author{Yimin Fu, Zhunga Liu, Dongxiu Guo, Longfei Wang
\thanks{
The authors are with the School of Automation, Northwestern Polytechnical University, Xi'an, China.
(e-mail: fuyimin96@mail.nwpu.edu.cn; liuzhunga@nwpu.edu.cn; gdx@mail.nwpu.edu.cn; ll\_wang@mail.nwpu.edu.cn).
}}

\markboth{Journal of \LaTeX\ Class Files}%
{Shell \MakeLowercase{\textit{et al.}}: A Sample Article Using IEEEtran.cls for IEEE Journals}
\maketitle

\begin{abstract} 
The acquisition of high-quality labeled synthetic aperture radar (SAR) data is challenging due to the demanding requirement for expert knowledge.
Consequently, the presence of unreliable noisy labels is unavoidable, which results in performance degradation of SAR automatic target recognition (ATR). 
Existing research on learning with noisy labels mainly focuses on image data. 
However, the non-intuitive visual characteristics of SAR data are insufficient to achieve noise-robust learning. 
To address this problem, we propose collaborative learning of scattering and deep features (CLSDF) for SAR ATR with noisy labels. 
Specifically, a multi-model feature fusion framework is designed to integrate scattering and deep features.
The attributed scattering centers~(ASCs) are treated as dynamic graph structure data, and the extracted physical characteristics effectively enrich the representation of deep image features.
Then, the samples with clean and noisy labels are divided by modeling the loss distribution with multiple class-wise Gaussian Mixture Models (GMMs).
Afterward, the semi-supervised learning of two divergent branches is conducted based on the data divided by each other.
Moreover, a joint distribution alignment strategy is introduced to enhance the reliability of co-guessed labels.
Extensive experiments have been done on the Moving and Stationary Target Acquisition and Recognition (MSTAR) dataset, and the results show that the proposed method can achieve state-of-the-art performance under different operating conditions with various label noises.
The code will be released at \url{https://github.com/fuyimin96/CLSDF} upon acceptance.
\end{abstract}  

\begin{IEEEkeywords}
Attributed scattering centers (ASCs), automatic target recognition (ATR), collaborative learning, noisy labels, synthetic aperture radar (SAR).
\end{IEEEkeywords}

\vspace{-0.2cm}
\section{Introduction}
\begin{figure}[]
\centering \vspace{-0.2cm}
\includegraphics[width=0.5\textwidth]{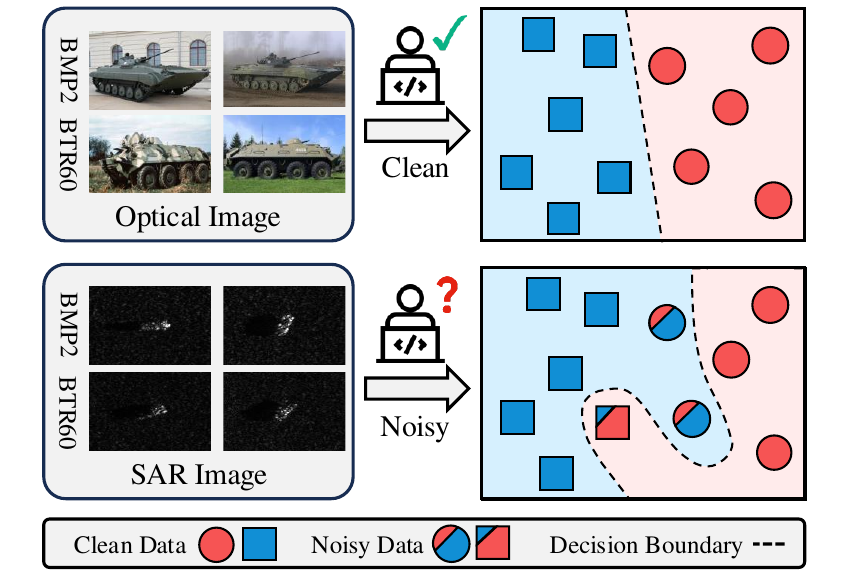}
\caption{Training the model on correctly labeled clean data of optical  images and mislabeled noisy data of SAR images.}
\label{Loss_curve}\vspace{-0.3cm}
\end{figure}

\IEEEPARstart{S}{YNTHETIC} aperture radar (SAR) serves as an active imaging system with all-time and all-weather perception ability~\cite{moreira2013tutorial,saeedi2017feasibility}.
Automatic target recognition (ATR)~\cite{bhanu1986automatic}, a crucial application for SAR image interpretation, is pivotal in both military and civilian domains.
Recent advancements in deep neural networks (DNNs)~\cite{he2016deep,karoly2020deep,fukushima2021artificial,fu2025reason,xu2025mambahsisr} have led to continuous improvements in SAR ATR, but its success is heavily dependent on the quality of data annotation.
However, the interpretation of SAR data is challenging due to the lack of visual intuitiveness.
During the annotation process, specialized expert knowledge is required, which is resource-intensive and impractical for large-scale datasets.
Moreover, collected SAR data often involve non-cooperative targets~\cite{schumacher2005non} without identifying signals, along with interference of geometric distortions and speckle noise.
Consequently, completely accurate annotation is hard to ensure in real-world SAR ATR applications, making model training inevitably subject to mislabeled samples—known as the noisy label problem~\cite{natarajan2013learning}.
As illustrated in Fig.~\ref{Loss_curve}, the model can effectively learn discriminative decision boundaries from correctly labeled clean data. 
In contrast, the decision boundaries tend to overfit to mislabeled noisy data, severely degrading generalization and recognition reliability.

Numerous research on learning with noisy labels~\cite{song2022learning} has been conducted to achieve noise-robust training of DNNs, which are mainly implemented from four perspectives: model architecture~\cite{sukhbaatar2014training,goldberger2016training,yao2018deep}, regularization technique~\cite{zhang2017mixup,jenni2018deep,menon2019can}, 
loss function~\cite{zhang2018generalized,wang2019symmetric,liu2015classification,patrini2017making}, and sample selection~\cite{han2018co,wei2020combating,li2020dividemix}.
Despite the widely validated effectiveness, existing methods are mainly focused on learning with noisy labels from single image data.
Since the visual characteristics of SAR data are non-intuitive, only using the deep image features is insufficient to overcome the negative impact of noisy labels.
Besides, the variants in depression angle and version under different operating conditions also decrease the recognition performance of deep learning-based methods.
Therefore, the generalization capability of DNNs cannot be guaranteed in the presence of noisy labels, and the representation of deep image features extracted by the convolutional neural network (CNN) need to be enriched with additional characteristics.

In addition to amplitude images, attributed scattering centers (ASCs)~\cite{potter1997attributed} represent SAR targets from another perspective, which contains inherent physical characteristics in the frequency domain.
Compared with deep image features, electromagnetic scattering features are relatively invariant and more robust to azimuth angle variations and resolution differences~\cite{zhou2014three}, enabling the stable representation of targets under complex operating conditions.
Recently, fusing diverse features has shown strong effectiveness across various remote sensing tasks~\cite{zhang2023feature,fu2024class,tao2025spectral}, and some fusion frameworks of scattering and deep features~\cite{zhang2020fec,liu2022multilevel,sunscan} have been proposed to improve SAR ATR performance.
However, how scattering and deep features can be integrated to facilitate the realization of noise-robust training remains challenging.
First, the representation inconsistency between different modalities will obstruct the fusion of scattering and deep features, causing incomplete utilization of complementary information.
Second,	the changes of operating conditions will bring additional disturbances to SAR ATR, making it difficult to stably avoid confirmation bias~\cite{tarvainen2017mean} with different types and levels of label noise. 

In this paper, we propose collaborative learning of scattering and deep features (CLSDF) for SAR ATR with noisy labels.
Different from existing image-based methods, CLSDF fully leverages the perception characteristics of SAR data to improve generalization capability against noisy labels.
Specifically, the multi-model fusion framework integrates scattering and deep features extracted from the perspective of graph structure and pixel grid, respectively.
Based on the enriched representation, the training process is conducted in a semi-supervised manner with two divergent branches.
First, the loss distributions of each class are modeled with multiple two-component Gaussian Mixture Models (GMMs)~\cite{pernkopf2005genetic,huang2017robust}.
Afterward, the separated samples with clean and noisy labels are mutually employed to train the other branch.
Then, the noisy labels are refurbished by label co-guessing to ensure the full exploration of all training data.
Moreover, the joint distribution of clean labels and refurbished noisy labels is encouraged to match the marginal class distribution, making the results of label co-guessing more reliable.
Finally, ensemble predictions of two branches are used to produce the final recognition results during testing.
The effectiveness of the proposed method has been thoroughly validated on the Moving and Stationary Target Acquisition and Recognition (MSTAR) dataset~\cite{keydel1996mstar}.
Compared to state-of-the-art methods, the proposed method can achieve advanced performance under different operating conditions across various types and levels of label noise.

Our main contributions can be summarized as follows:
\begin{enumerate}
\item We propose a collaborative learning method that integrates scattering and deep features for SAR ATR with noisy labels. To the best of our knowledge, this is the first time that physical and visual characteristics are jointly used to achieve noise-robust training.
\item We propose a multi-model feature fusion framework to extract and fuse features from diverse perspectives. 
Based on the enriched representation, the loss distribution is modeled by multiple class-wise GMMs, enabling more accurate separation of clean and noisy labels. 
\item We propose a joint distribution alignment strategy to facilitate the semi-supervised learning process. 
By constraining the joint distribution of clean and refurbished noisy labels, the accumulation of unreliable label guessing results can be adaptively calibrated.
\end{enumerate}

The rest of this paper is organized as follows.
Section \ref{sec2} provides a comprehensive review of the related research progress.
In Section \ref{sec3}, technical details of the proposed method are elaborated.
Experimental validation and performance analysis are carried out in Section \ref{sec4}.
Finally, we conclude the work in this paper and prospect future research in Section \ref{sec5}.

\begin{figure*}[]
\centering \vspace{-0.7cm}
\subfloat[20\% symmetric noise]{\includegraphics[width=0.3\textwidth,height=0.21\textwidth]{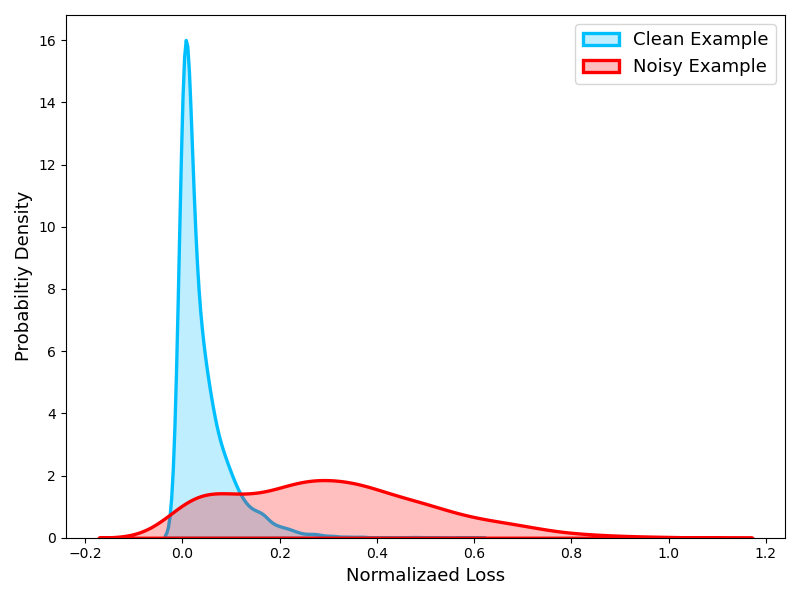}\label{sym2}}\hspace{0.3cm}
\subfloat[20\% asymmetric noise ]{\includegraphics[width=0.3\textwidth,height=0.21\textwidth]{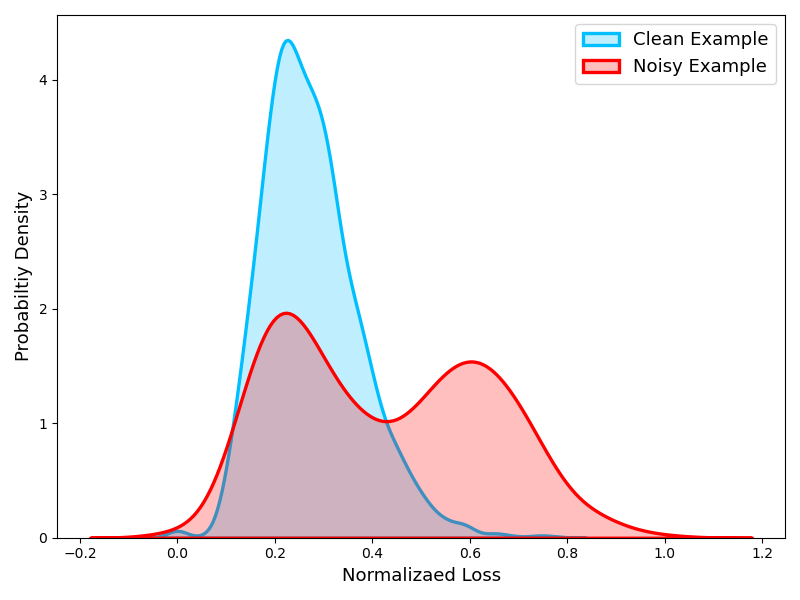}\label{asym2}}\hspace{0.3cm}
\subfloat[80\% symmetric noise]{\includegraphics[width=0.3\textwidth,height=0.21\textwidth]{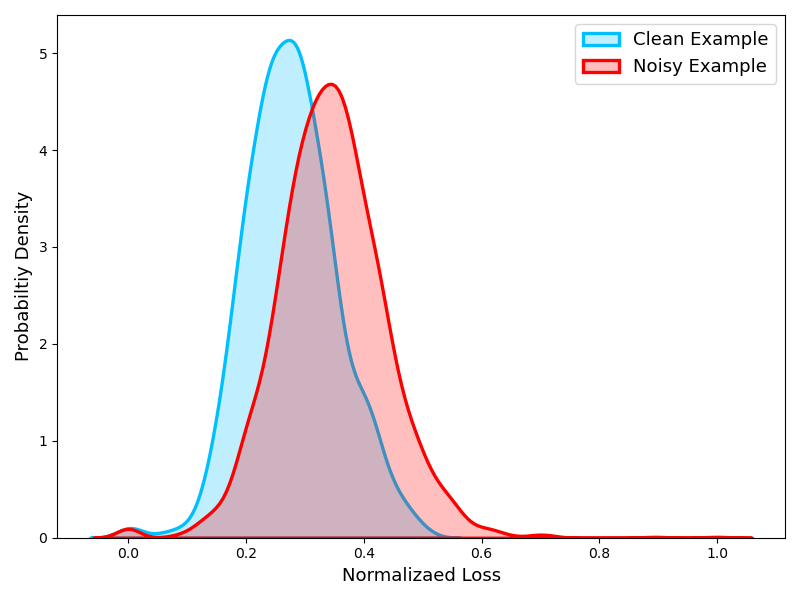}\label{sym8}}
\caption{Probability densities of normalized loss of samples with clean and noisy labels under different settings of noises.}
\label{loss_dis} \vspace{-0.3cm}
\end{figure*}

\section{Related Works}\label{sec2}
In this section, we first introduce the related works of learning with noisy labels. 
Then, we give a brief review of recent progress in SAR target recognition.

\subsection{Learning with Noisy Labels}
Existing methods of learning with noisy labels can be divided into four types based on the approaches to mitigate overfitting to mislabeled samples: 

\textit{1) Model architecture-based methods.} 
This type of method usually adopts a noise adaptation layer or specialized architecture to estimate the noise transition matrix.
Sukhbaatar et al.~\cite{sukhbaatar2014training} imposed a constrained noise layer on DNNs to learn the distribution of noisy labels.
Goldberger and Ben-Reuven~\cite{goldberger2016training} applied the expectation–maximization (EM) algorithm to predict the correct labels of mislabeled samples by an additional softmax layer.
To prevent errors caused by misestimated noise transition, Yao et al.~\cite{yao2018deep} proposed a contrastive-additive noise network (CAN), which can measure the trustworthiness of noisy labels by quality embedding.
However, the estimation accuracy of the transition matrix and the extension flexibility of the model architecture cannot be satisfied simultaneously.

\textit{2) Regularization technique-based methods.}
The tolerance to noisy labels can be effectively improved by regularized DNNs explicitly and implicitly.
Zhang et al.~\cite{zhang2017mixup} proposed mixup to improve the generalization of DNNs with linear interpolation-based augmented data.
Jenni and Favaro~\cite{jenni2018deep} proposed a bilevel learning strategy to regularize the training process by a validation set.
Menon et al.~\cite{menon2019can} proposed composite loss-based gradient clipping as an noise-robust modification to underlying loss functions.
Liu et al.~\cite{liu2020early} proposed early-learning regularization (ELR) to eliminate the interference of mislabeled samples on the learning of classification.
Despite the simplicity of implementation, the effects of regularization techniques are usually sensitive to hyperparameter settings.

\textit{3) Loss function-based methods.}
A theoretically intuitive approach to reduce the negative impact of noisy labels is to train the network with a noise-robust loss function.
Zhang and Sabuncu~\cite{zhang2018generalized} generalized cross entropy (CE) loss by mean absolute error (MAE) loss~\cite{ghosh2017robust} for noise-robust classification.
Wang et al.~\cite{wang2019symmetric} proposed symmetric cross entropy (SCE) loss to simultaneously solve the under learning and overfitting problems.
Apart from designing noise-robust loss functions, adjusting the loss of training samples during the optimization process also helps.
Liu and Tao~\cite{liu2015classification} proposed an importance reweighting framework with the optimization priority to noise-free samples.
Patrini et al.~\cite{patrini2017making} proposed two loss correction procedures based on linear combinations of loss values and network predictions.
Arazo et al.~\cite{arazo2019unsupervised} corrected the loss based on the convex combination of network predictions and noisy labels.
Nevertheless, the performance of loss function-based methods can be compromised in complex scenarios.

\textit{4) Sample selection-based methods.}
Compared with other types of methods, the sample selection-based methods are more effective in avoiding confirmation bias, which also enables the full exploration of training data in conjunction with semi-supervised learning~\cite{mey2022improved}.
Malach and Shalev-Shwartz~\cite{malach2017decoupling} proposed decoupling to update two networks based on their disagreement.
Inspired by the memorization effect~\cite{arpit2017closer} of DNNs, Han et al.~\cite{han2018co} proposed the co-teaching learning framework, where the clean labels selected by two DNNs are used to teach the peer network mutually.
To avoid the over consensus between two DNNs, Yu et al.~\cite{yu2019does} employed a disagreement-based update strategy into training.
Wei et al.~\cite{wei2020combating} proposed a joint training method with co-regularization (JoCoR) from the perspective of agreement maximization.
Tan et al.~\cite{tan2021co} implemented different learning strategies in two branch networks to provide new perspectives on noise-robust training.
Karim et al.~\cite{karim2022unicon} combined uniform selection and contrastive learning (UNICON) to address the class imbalance among the selected clean subset.
To leverage samples with both clean and noisy labels, Li et al.~\cite{li2020dividemix} proposed DivideMix to train the model in a semi-supervised manner.
Based on a two-component GMM, the training data can be split into labeled and unlabeled sets, on which label co-refinement and co-guessing are preformed to train the other network.
Although the robustness to noisy labels is significantly improved, the sample selection-based methods are susceptible to changes in data and noise, which is inevitable during the recognition process of SAR targets.

While these methods have demonstrated strong performance for natural image classification with noisy labels, their direct applicability to SAR ATR is limited due to the lack of visual intuitiveness.
Some investigations have been conducted to alleviate the negative impact of noisy labels in SAR data on scene classification~\cite{huangclassification}, object detection~\cite{pillaideepsar}, and change detection~\cite{mengsynthetic}.
Recently, a loss curve-fitting-based method~\cite{wang2021label} has been proposed for SAR ATR with noisy labels, but still only from the perspective of visual images without considering the inherent perceptual characteristics of SAR data.
In addition, the performance evaluation is conducted solely under standard
operating conditions (SOCs) with symmetric noise, which is incomprehensive.
As shown in Fig.~\ref{loss_dis}, the complexity of sample selection increases considerably as the noise rate increases and the type of noise changes. 
Consequently, a comprehensive exploration of SAT ATR in the presence of noisy labels is needed.
\subsection{SAR Target Recognition}
In addition to conventional geometric~\cite{akbarizadeh2012new} and transform domain features~\cite{pei2016sar}, existing SAR target recognition methods are mainly based on two types of features:

\textit{1) Deep image features.}
Benefiting from the outstanding feature extraction ability of DNNs, significant improvements have been achieved based on deep image features.
Chen et al.~\cite{chen2016target} proposed all-convolutional networks (A-ConvNets) as the first deployment of data-driven features for SAR ATR.
Shang et al.~\cite{shang2018sar} used an information recorder to extract spatial similarity information to improve the recognition performance.
Zhou et al.~\cite{zhou2018sar} designed a large-margin softmax batch-normalization structure to improve the generalization performance of the convolutional neural network (CNN).
To improve SAT ATR performance with limited training samples, Pei et al.~\cite{pei2017sar} proposed a multiview CNN based on the parallel topology structure. 
Wen et al.\cite{wen2021rotation} developed a self-supervised learning framework that incorporates rotation awareness.
However, the representation provided by deep image features is limited under complex operating conditions, resulting in severe degradation of recognition performance.

\textit{2) Electromagnetic scattering features.}
Different from deep image features, electromagnetic scattering features are robust to variations in azimuth angle, resolution and signal-to-noise ratio, and can steady reflect the inherent physical characteristic of the targets.
Zhou et al.~\cite{jianxiong2011automatic} established a global
scattering center model based on range profiles at multiple viewing angles for SAR ATR.
Fu et al.~\cite{fu2018aircraft} matched the scattering structure feature (SSF)  to improve aircraft recognition accuracy in SAR images.
Ding et al.~\cite{ding2018efficient} hierarchically fused random projection and scattering features to combine the advantages of global and local descriptors.
Huang et al.~\cite{huang2022physically} injected  scattering features into DNNs to enhance the explainability for SAR image classification.

Considering the respective advantages of scattering and deep features in different scenarios, some fusion frameworks have been proposed.
Jiang et al.~\cite{jiang2018hierarchical} designed a reliability-based fusion strategy of CNN and ASC for hierarchical classification.
Zhang et al.~\cite{zhang2020fec} fused ASC and CNN features at the feature level based on discrimination correlation analysis (DCA).
To balance recognition accuracy and generalization, Feng et al.~\cite{feng2021sar} combined local and global features extracted by the ASC part model and all-convolutional network.
Liu et al.~\cite{liu2022multilevel} regarded ASC as set-data to extract multilevel scattering feature, which are fused with deep features by a scattering and deep feature fusion network (SDF-Net).
Li et al.~\cite{li2022novel} integrated scattering center extraction with graph construction for efficient ATR with limited data.
Sun et al.~\cite{sunscan} combined scattering extraction with the imaging mechanism in the meta-learning process for few-shot aircraft recognition.
Feng et al.~\cite{fengelectromagnetic} integrated the electromagnetic characteristics into deep learning features to improve the recognition stability under complex conditions.
In addition to vehicle recognition, Zhang et al.~\cite{zhang2023scattering} incorporated scattering features into the learning process of DNN for few-shot SAR ship classification.	
Zhao et al.~\cite{zhao2023scattering} proposed a spatial-structural association framework for effectively integrating scattering and image features of aircraft.
However, existing methods typically handle different scattering centers individually or model the spatial relations based on static graphs.
Consequently, the physical characteristics of the target cannot be fully explored due to the insufficient capture of topological information~\cite{10226282}. 
Besides, there is a void in how scattering and deep features can be jointly used to improve the robustness to label noise.

\section{Methodology}\label{sec3} 
In this section, we first provide preliminary of how to construct ASC model from the SAR data.
Afterward, we overview the architecture of CLSDF and how noise-robust training is achieved.
Then, each crucial part of the proposed method is introduced in detail.

\subsection{Preliminary of Parametric ASC Model}
Motivated by physical optics and geometric theory of diffraction (GTD)~\cite{keller1962geometrical}, the parametric ASC model~\cite{gerry1999parametric} describes scattering centers of the target as a function of frequency and aspect angle, which have been widely used in target recognition.
The total scattered field can be represented by the summarized responses of $P$ individual scattering centers:
\begin{equation}
E(f, \varphi ; \Theta)=\sum_{i=1}^P E_i\left(f, \varphi ; \theta_i\right),
\end{equation}
where $f$ and $\varphi$ represent the operating frequency and aspect angle of the radar, respectively.
Specifically, the scattering field of each scattering center $i$ can be expressed as:
\begin{equation}
\begin{aligned}
E_i\left(f, \varphi ; \theta_i\right)= & A_i \cdot\left(j \frac{f}{f_c}\right)^{\alpha_i} \\
& \cdot \exp \left[-j \frac{4 \pi f}{C}\left(x_i \cos \varphi+y_i \sin \varphi\right)\right] \\
& \cdot \operatorname{sinc}\left(\frac{2 \pi f}{C} L_i \sin \left(\varphi-\bar{\varphi}_i\right)\right) \\
& \cdot \exp \left(-2 \pi f \gamma_i \sin \varphi\right),
\end{aligned}
\end{equation}
where $C$ and $f_c$ refer to the propagation speed of electromagnetic wave and center frequency of radar, respectively.
The physical parameter set $\theta_i$ contains different physical descriptions of the target:
\begin{equation}
\theta_i = \left[A_i, x_i, y_i, \alpha_i, L_i, \bar{\phi}_i, \gamma_i\right],
\end{equation}
where $A_i$ is a scalar of the amplitude, and $(x_i, y_i)$ is the position vector. 
$\alpha_i$ represents the geometry dependence, $L_i$ and $\bar{\phi}_i$ refer to the length and direction of the distributed scattering center.
The value of $\gamma_i$ is usually small, which indicates the aspect dependence of the distributed scattering center.
The estimation of parameter set $\Theta$ for all scattering centers is carried out based on the algorithm proposed in our previous work~\cite{liu2022multilevel}, where the number of estimated scattering centers is set as a constant $P$.
It should be noted that the actual number of scattering centers may differ from $P$.
If the actual number is less than $P$, extra points from background clutter or redundant target regions will be selected, including irrelevant information.
On the contrary, some critical information may be lost.
Since the setting of $P$ is crucial, we conduct experiments in Subsection~\ref{hyper_ana} to determine an appropriate value.

\begin{figure*}[]
\centering \vspace{-0.5cm}
\includegraphics[width=0.9\textwidth]{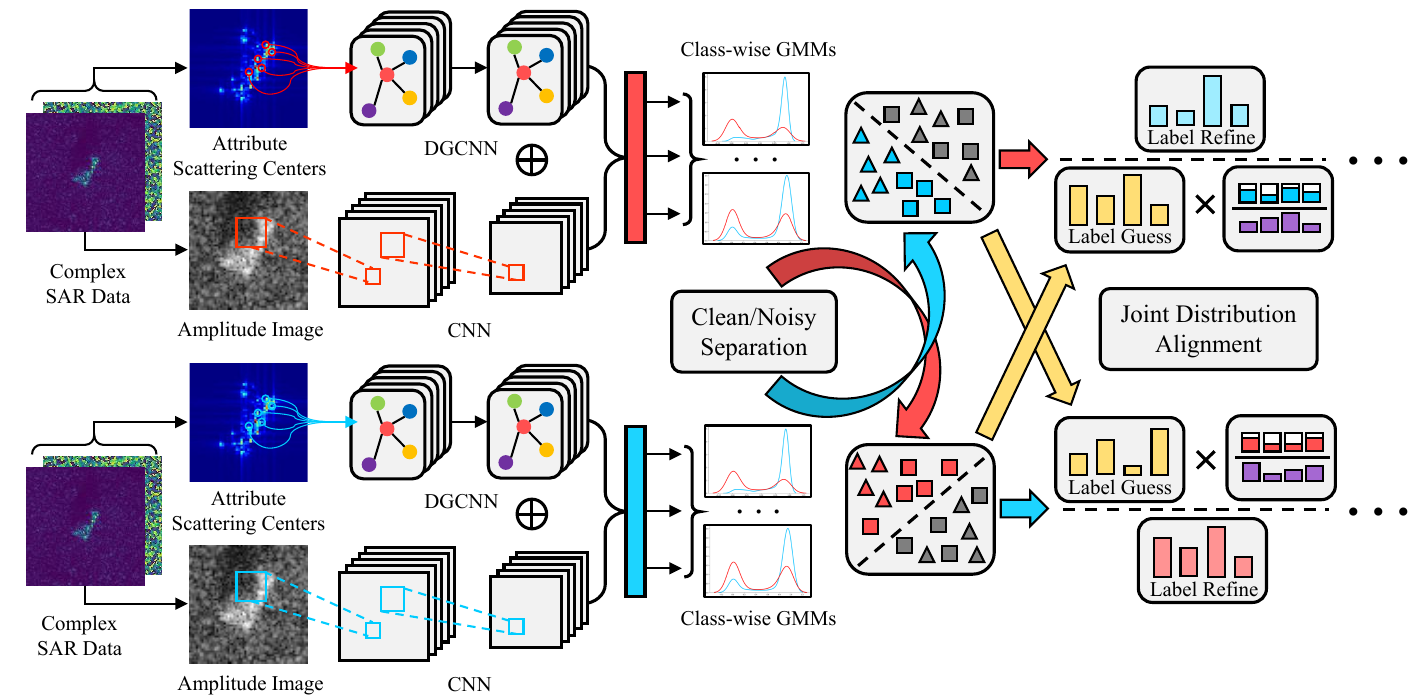}
\caption{Overview of the proposed CLSDF with three main parts: multi-model feature extraction, class-wise sample selection, and joint distribution alignment.}
\label{model} \vspace{-0.25cm}
\end{figure*}

\subsection{Architecture Overview}
The architecture overview of the proposed collaborative learning of scattering and deep features is shown in Fig.~\ref{model}.
During training, two divergent network branches are simultaneously trained in a semi-supervised manner to avoid confirmation bias.
First, the ASC and amplitude images are input into the multi-model feature fusion framework of each branch. 
Afterward, the scattering features are extracted based on a local neighborhood graph and fused with the deep image features.
Then, multiple class-wise GMMs are fitted to separate samples with clean and noisy labels, which are used to train the other branch.
During the semi-supervised learning process, the refurbishment of noisy labels is calibrated by the joint distribution alignment strategy.
Finally, the ensemble predictions of the two branches are used as the final prediction during testing.
In the following subsections, the technical details of each crucial part will be interpreted.

\subsection{Multi-model Feature Extraction} 
The lack of visual intuitiveness in SAR data makes the integration with physical characteristics indispensable to overcome noisy labels.
Since the effectiveness of CNN for image feature extraction has been widely validated, the core problem of representation enrichment becomes the extraction of scattering features.
However, existing methods usually treat ASC as separate individual points but ignore the topological relations between different scattering centers, leading to insufficient representation of inherent physical characteristics.
In comparison, dynamic graph CNN (DGCNN)~\cite{wangdynamic} employs edge convolution along the edges of a K-nearest neighbor (KNN) graph to extract edge features rather than individual point features, which enables effective capture of geometric and topological relationships among points.	
Besides, the dynamic update of the local neighborhood graph allows for layer-wise adaptation to evolving feature representations, thereby supporting the exploration of semantic information from long-range dependencies.
Moreover, CNN and DGCNN can extract features from different perspectives, whose joint use can facilitate the exploration of complementary information.

Therefore, we design a multi-model feature fusion framework to extract scattering and deep features, and take fully advantage of them by feature concatenation.
Giving the training dataset $\mathcal{D}=(\mathcal{X},\mathcal{Y})=\{(x_n,y_n)\}_{n=1}^N$ with N samples, where $x_n$ and $y_n$ represent the complex SAR data and the class label of the target, respectively.
For each sample, the complex SAR data $x$ can be converted into the combination of a set of ASC $x_S \in \mathbb{R}^{P \times 7}$ and an amplitude image $x_I\in~\mathbb{R}^{H \times W}$, where $H \times W$ is the resolution of the amplitude image.

First, the deep feature $z_I\in \mathbb{R}^{d_I}$ can be extracted from the amplitude image through stacked convolutional blocks and a global average pooling (GAP) layer:
\begin{equation}
z_I=\text{GAP}(\text{Conv}(x_I)).
\end{equation}
Then, we initialize a local neighborhood graph $\mathcal{G}$ by setting each scattering center as the vertexes.
The edge relationships of each vertex is constructed by connecting with its KNNs, and the edge features $e_i$ can be represented as:
\begin{equation}
e_i=\{e_{i1},\ldots,e_{iK}\},~i\in \{1,\ldots,P\}
\end{equation}
\begin{equation}
e_{ij}=\mathcal{H}\left(x_{S_i}, x_{S_j}-x_{S_i}\right),~j\in \{1,\ldots,K\}
\end{equation}
where $\mathcal{H}(\cdot)$ represents a learnable nonlinear transformation, employing a two-dimensional convolutional layer with a leaky rectified linear unit here.
Afterward, the edge features associated with different edges are aggregated by a channel-wise sum operation, which are used as the representation of the corresponding vertex in the next graph layer:
\begin{equation}
x_{S_i}^1=\sum_{j=1}^K~\mathcal{H}\left(x_{S_i}, x_{S_j}-x_{S_i}\right),
\end{equation}
which can provide more neighborhood information.
In subsequent layers, the KNNs of each vertex need to be redetermined according to the pairwise distance in the feature space.
When finishing the dynamic update of L-layer graph structure, the features of each layer $l \in \{1,\ldots,L\}$ are concatenated to aggregate multi-scale information.
Moreover, global max pooling (GMP) and GAP operations are separately applied to the concatenated representations.
To comprehensively explore the physical characteristics, the outputs of the above two pooling operations are integrated into scattering features $z_S$:
\begin{equation}
z_S = \text{GMP}(x_{S}^1\oplus x_{S}^2\ldots\oplus x_{S}^L) \oplus \text{GAP}(x_{S}^1\oplus x_{S}^2\ldots\oplus x_{S}^L),
\end{equation}
where $x_{S}^l=\{x_{S_1}^l,\ldots,x_{S_P}^l\}$ and $\oplus$ represents the concatenation operation.
Finally, the deep and scattering features is fused into $z_F$:
\begin{equation} \label{eq_fusion}
z_F = z_S \oplus z_I,
\end{equation}	
which provides a multi-characteristic representation of the SAR target from different perspectives.

\begin{figure}[]
\centering \vspace{-0.4cm}
\subfloat[T72]{\includegraphics[width=0.25\textwidth]{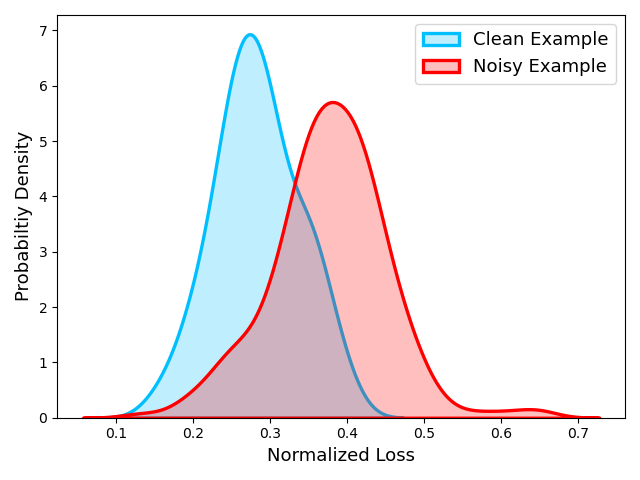}\label{class1}}
\subfloat[ZIL131]{\includegraphics[width=0.25\textwidth]{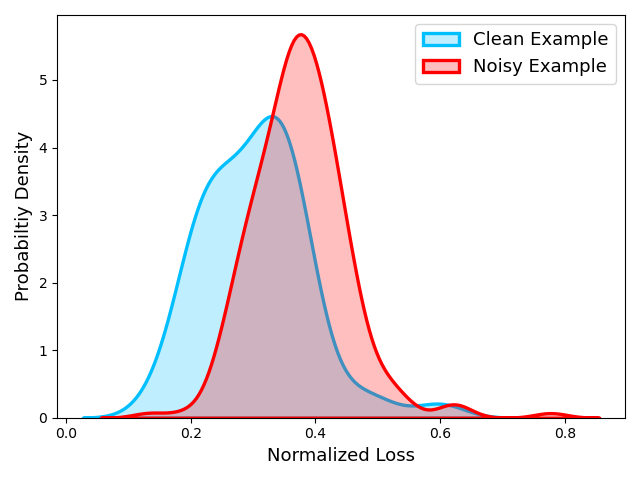}\label{class2}}
\caption{Comparison of per-sample loss distribution between different classes.}
\label{class_wise} \vspace{-0.3cm}
\end{figure}

\subsection{Class-wise Sample Selection}
After integrating scattering and deep features of the target, each branch is supposed to divide the training samples into clean and noisy subsets, which are used to conduct semi-supervised learning of the other branch.
Therefore, the efficacy of subsequent learning processes depends heavily on the reliability of data division.
Since clean samples tend to be learned faster by DNNs than noisy samples~\cite{arpit2017closer}, existing methods~\cite{arazo2019unsupervised,li2020dividemix} have attempted to model the per-sample loss distribution with a single mixture model for data division.	
Nonetheless, as shown in Fig.~\ref{class_wise}, the per-sample loss distribution exhibits inconsistency across different categories.
Consequently, the data division based on a single class-independent mixture model is imprecise.

To solve this problem, we adopt multiple class-wise GMMs to select samples of each class into clean and noisy subsets.
First, the per-sample loss of each class is calculated as:
\begin{equation}
\mathcal{L}_c = \{l_n\}_{{y_n}=c}=\{-y_n \log \left(\ell\left(z_{F_n}\right)\right)\}_{{y_n}=c},~c\in\{1, \ldots, C\}
\end{equation}	
where $C$ is the total number of classes and $\ell(\cdot)$ is the classification function.
Then, the loss distribution of each class is used to fit a two-component GMM.
For each sample within the corresponding class, the probability $\pi_n$ that it is correctly labeled can be derived from the density of the Gaussian component with a smaller mean on it.
Afterward, the data of each class can be divided based on a probability threshold~$\delta$. 
If the probability of being correctly labeled exceeds this threshold, this sample will be selected into the clean subset and keep its original class label.
Otherwise, it will be regarded as a mislabeled sample whose original class label will be removed:
\begin{equation}
(x_n,y_n)= \begin{cases}(x_n,y_n,\pi_n)\rightarrow \mathcal{D}_{clean}, & \text { if } \pi_n \geq \delta \\ 
x_n\rightarrow \mathcal{D}_{noisy},& \text { otherwise }\end{cases}
\end{equation}
where $\mathcal{D}_{clean}$ and $\mathcal{D}_{noisy}$ represent clean and noisy subsets, respectively.
Finally, the divided data $\tilde{\mathcal{D}}=\{\mathcal{D}_{clean},\mathcal{D}_{noisy}\}$
from each branch are used to conduct semi-supervised learning of the conjugate branch.

\subsection{Semi-supervised Learning with Joint Distribution Alignment}
With the interchanged data division, two branches are trained alternately in epochs. 
In addition to supervised learning on samples in the clean subset, the non-labeled samples in the noisy subset are also leveraged via semi-supervised learning. 
To ensure comprehensive exploration of the training data, the guessed/generated pseudo-labels are usually assigned to non-labeled samples.
However, the label guessing results are prone to over-accumulate on certain classes due to the confirmation bias~\cite{arazo2020pseudo}.
As a result, the semi-supervised learning can be severely affected by those unreliable relabeled samples.

To obviate this problem and achieve noise-robust training, we propose a joint distribution alignment strategy to improve the reliability of label guessing.
First, label co-refinement for samples in the clean subset is performed following~\cite{li2020dividemix}, which combines the original label $y_n$ with the average prediction of multiple augmentations of the input:
\begin{equation}
\bar{y}_n=\pi_n y_n+\left(1-\pi_n\right)\frac{1}{M} \sum_{m=1}^M\ell\left(z_{F_n}^m\right),
\end{equation}
where $M$ is the number of augmentations, and $z_{F_n}^m$ is the extracted fusion feature of the corresponding augmented sample~$\hat{x}_n^m$.
Next, the predictions from each branch are averaged as the initial guessing labels of samples in the noisy subset, i.e., label co-guessing:
\begin{equation}
{q}_n=\frac{1}{2M} \sum_{m=1}^M(\ell\left(z_{F_n}^m\right)+{\ell'}\left({z}_{F_n}^m\right)),
\end{equation}
where ${\ell'}(\cdot)$ represents the classification function of the other branch.
Then, we calibrate the guessing labels by matching the joint distribution of clean labels and initial guessing results to the marginal class distribution $p(y)$ of the non-prior prediction:
\begin{equation}
\bar{q}_n=\text{Norm}({q}_n\times \frac{p(y)-\tilde{p}(y)}{\tilde{p}(q)}),
\end{equation}
where $\tilde{p}(y)$ and $\tilde{p}(q)$ represent the distributions of clean labels versus overall data and the accumulative average predictions versus relabeled samples, respectively.
Afterward, a temperature sharpening function is applied on both refined and guessed labels to minimize the entropy of distribution:
\begin{equation}
\begin{cases}
\hat{y}_n=\operatorname{Sharpen}(\bar{y}_n, T)={\bar{y}_n}^{\frac{1}{T}} / \sum_{c=1}^C {\bar{y}_n}^{c \frac{1}{T}},\\
\hat{q}_n=\operatorname{Sharpen}(\bar{q}_n, T)={\bar{q}_n}^{\frac{1}{T}} / \sum_{c=1}^C {\bar{q}_n}^{c \frac{1}{T}}.
\end{cases}
\end{equation}
Following MixMatch~\cite{berthelot2019mixmatch}, the refined clean subset $\bar{\mathcal{D}}_{clean}$ and relabeled noisy subset $\bar{\mathcal{D}}_{noisy}$ are combined, from which the randomly selected sample pairs $\{(x_1,p_1),(x_2,p_2)\}$ are `mixed' by linear interpolation:
\begin{equation}
\begin{aligned}
\begin{cases}
\lambda & \sim \operatorname{Beta}(\alpha, \alpha), \\
\lambda^{\prime} & =\max (\lambda, 1-\lambda), \\
x^{\prime} &=\lambda^{\prime} x_1+\left(1-\lambda^{\prime}\right) x_2, \\
p^{\prime} & =\lambda^{\prime} p_1+\left(1-\lambda^{\prime}\right) p_2 .
\end{cases}
\end{aligned}
\end{equation}
where $p_1$ and $p_2$ represent the probabilistic labels. 
Therefore, the data from clean and noisy subsets is converted into ${\mathcal{D}}_{clean}^{\prime}$ and ${\mathcal{D}}_{noisy}^{\prime}$.
Finally, the network branch is semi-supervised by the cross-entropy loss on ${\mathcal{D}}_{clean}^{\prime}$ and the mean squared error loss on ${\mathcal{D}}_{noisy}^{\prime}$:
\begin{equation}
\mathcal{L}_{CE}  =\frac{1}{\left|{\mathcal{D}}_{clean}^{\prime}\right|} \sum_{x^{\prime}, p^{\prime} \in {\mathcal{D}}_{clean}^{\prime}} -p^{\prime} \log \left(\ell\left(z_{F}^{\prime}\right)\right),
\end{equation}
\begin{equation}
\mathcal{L}_{MSE}  =\frac{1}{\left|{\mathcal{D}}_{noisy}^{\prime}\right|} \sum_{x^{\prime}, p^{\prime} \in {\mathcal{D}}_{noisy}^{\prime}}\left\|p^{\prime}-\ell\left(z_{F}^{\prime}\right)\right\|_2^2,
\end{equation}
\begin{equation}
\mathcal{L}  =\mathcal{L}_{CE}+\lambda_u \mathcal{L}_{MSE},
\end{equation}
where $\lambda_u$ is a hyperparameter to control the weight of mean squared error loss.

\section{Experiments}\label{sec4}
\begin{figure}[] \vspace{-0.4cm}
\centering
\includegraphics[width=0.5\textwidth]{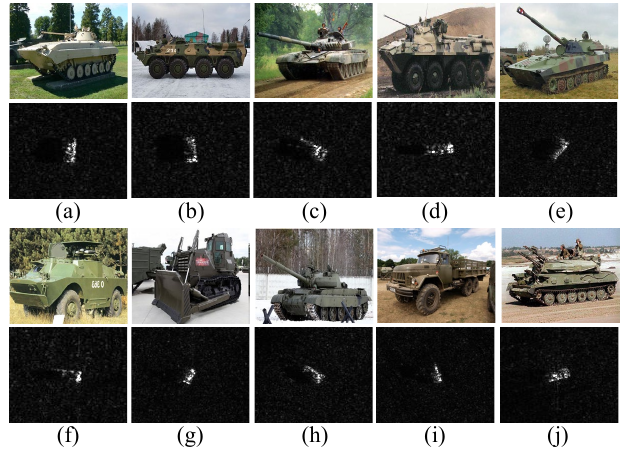} \vspace{-0.5cm}
\caption{The corresponding optical and SAR images of each class in the MSTAR dataset.
(a) BMP2. (b) BTR70. (c) T72. (d) BTR60. (e) 2S1. (f) BRDM2.
(g) D7. (h) T62. (i) ZIL131. (j) ZSU234.} \vspace{-0.1cm}
\label{mstar} \vspace{-0.2cm}
\end{figure}

\subsection{Dataset Description} 
To evaluate the effectiveness of our proposed method for SAR ATR with noisy labels, extensive experiments have been done on the MSTAR data set. 
The SAR data of the MSTAR dataset is collected by the Sandia National Laboratory SAR sensor platform at the X-band HH polarization across the full range of viewing angles.
There are ten types of military vehicles contained in the MSTAR dataset,
and examples of optical and SAR images for each class are shown in Fig.~\ref{mstar}.
The operating conditions can be divided into the standard operating condition (SOC) and various extended operating conditions (EOCs) based on different observation configurations.
The adopted EOCs involve variations in depression angle and target version, as well as noise corruption.
Data corresponding to the first two variations are officially provided by the MSTAR dataset, while noise corruption is typically simulated using Gaussian noise in existing research~\cite{zhang2020fec,chen2016target,jianxiong2011automatic,ding2018efficient}.

\subsubsection{Data descriptions of SOC}
The training and test data under SOC is acquired under similar imaging conditions with proximate depression angles ($17^{\circ}$ for the training data and $15^{\circ}$ for the test data).
The serial number and sample amount of ten classes are enumerated in Table~\ref{data_soc}.

\begin{table}[t] 
\caption{Information of Training and Test Data Under Soc.} 
\label{data_soc}  \vspace{-0.2cm}
\centering
\setlength{\tabcolsep}{4.5pt}
\renewcommand\arraystretch{1}
\begin{tabular}{|c|c|c|c|c|c|}
\hline \multirow{2}{*}{ Class }& \multirow{2}{*}{$\begin{array}{c}\text { Serial } \\ \text { Number }\end{array}$} & \multicolumn{2}{c|}{Training Set} & \multicolumn{2}{c|}{Test Set} \\ \cline{3-6}
& &Depression & Number & Depression & Number \\
\hline BMP2 & 9563 & $17^{\circ}$ & 233 & $15^{\circ}$ & 195 \\
\hline BTR70 & C71 & $17^{\circ}$ & 233 & $15^{\circ}$ & 196 \\
\hline T72 & 132 & $17^{\circ}$ & 232 & $15^{\circ}$ & 196 \\
\hline T62 & A51 & $17^{\circ}$ & 299 & $15^{\circ}$ & 273 \\
\hline BRDM2 & E-71 & $17^{\circ}$ & 298 & $15^{\circ}$ & 274 \\
\hline BTR60 & 7532 & $17^{\circ}$ & 256 & $15^{\circ}$ & 195 \\
\hline ZSU234 & D08 & $17^{\circ}$ & 299 & $15^{\circ}$ & 274 \\
\hline D7 & 13015 & $17^{\circ}$ & 299 & $15^{\circ}$ & 274 \\
\hline ZIL131 & E12 & $17^{\circ}$ & 299 & $15^{\circ}$ & 274 \\
\hline 2S1 & B01 & $17^{\circ}$ & 299 & $15^{\circ}$ & 274 \\
\hline
\end{tabular} \vspace{-0.2cm}
\end{table}

\subsubsection{Data descriptions of EOC-1 (four large depression variant)}
Three types of SAR target are observed at different depression angles: $17^{\circ}$ for the training data and $15^{\circ}$, $30^{\circ}$, $45^{\circ}$ for the test data.
The data distribution is shown in Table~\ref{data_eoc1}.
As SAR imaging is sensitive to changes in depression angles, it is important to maintain robust performance in this case.

\textsc{\begin{table}[] \vspace{-0.3cm}
\caption{Information of Training and Test Data Under EOC-1.} 
\label{data_eoc1}\vspace{-0.2cm}
\centering
\renewcommand\arraystretch{1}
\begin{tabular}{|c|c|c|c|c|c|}
\hline \multirow{3}{*}{ Class }& \multirow{3}{*}{$\begin{array}{c}\text { Serial } \\ \text { Number }\end{array}$}  &\multicolumn{4}{c|}{Depression}\\ \cline{3-6}
&&\multicolumn{1}{c|}{Training Set} & \multicolumn{3}{c|}{Test Set} \\ \cline{3-6}
& &$17^{\circ}$ & $15^{\circ}$ & $30^{\circ}$ & $45^{\circ}$  \\
\hline BRDM2 & E-71 & 298& 274 &287 & 303  \\
\hline ZSU234 & D08 &299 & 274 &288 & 303  \\
\hline 2S1 & B01& 299& 274 & 288 &303  \\
\hline
\end{tabular}
\end{table}}

\begin{table}[t] \vspace{-0.3cm}
\caption{Information of Training and Test Data Under EOC-2.} 
\label{data_eoc2}\vspace{-0.2cm}
\centering
\setlength{\tabcolsep}{5pt}
\renewcommand\arraystretch{1}
\begin{tabular}{|c|c|c|c|c|c|}
\hline \multirow{2}{*}{ Class }& \multirow{2}{*}{$\begin{array}{c}\text { Serial } \\ \text { Number }\end{array}$} & \multicolumn{2}{c|}{Training Set} & \multicolumn{2}{c|}{Test Set} \\ \cline{3-6}
& &Depression & Number & Depression & Number \\
\hline T72 & A64 & $17^{\circ}$ & 299 & $30^{\circ}$ & 288 \\
\hline BRDM2 & E-71 & $17^{\circ}$ & 298 & $30^{\circ}$ & 287 \\
\hline ZSU234 & D08 & $17^{\circ}$ & 299 & $30^{\circ}$ & 288 \\
\hline 2S1 & B01& $17^{\circ}$ & 299 & $30^{\circ}$ & 288 \\
\hline
\end{tabular} \vspace{-0.2cm}
\end{table}

\begin{table}[] \vspace{-0.2cm}
\caption{Information of Training and Test Data Under EOC-3.} 
\label{data_eoc3}\vspace{-0.2cm}
\centering
\renewcommand\arraystretch{1}
\begin{tabular}{|c|c|c|c|c|c|}
\hline \multirow{3}{*}{ Class } & \multirow{3}{*}{$\begin{array}{c}\text { Serial } \\
\text { Number }\end{array}$} & \multicolumn{4}{c|}{ Depression } \\
\cline { 3 - 6 } & & Training Set& \multicolumn{3}{c|}{Test Set} \\
\cline { 3 - 6 } & & $17^{\circ}$ & $15^{\circ}$ & $17^{\circ}$ & Total \\
\hline \multirow{3}{*}{ BMP2 } & 9563 & 233 & - & - & - \\
\cline { 2 - 6 } & 9566 & - & 196 & 232 & 428 \\
\cline { 2 - 6 } & C21 & - & 196 & 233 & 429 \\
\hline BRDM2 & E-71 & 298 & - & - & - \\
\hline BTR70 & C71 & 233 & - & - & - \\
\hline \multirow{6}{*}{ T72 } & 132 & 232 & - & - & - \\
\cline { 2 - 6 } & 812 & - & 195 & 231 & 426 \\
\cline { 2 - 6 } & $\mathrm{A} 04$ & - & 274 & 299 & 573 \\
\cline { 2 - 6 } & $\mathrm{A} 05$ & - & 274 & 299 & 573 \\
\cline { 2 - 6 } & $\mathrm{A} 07$ & - & 274 & 299 & 573 \\
\cline { 2 - 6 } & $\mathrm{A} 10$ & - & 271 & 296 & 567 \\ 
\hline
\end{tabular}  \vspace{-0.2cm}
\end{table}

\subsubsection{Data descriptions of EOC-2 (two large depression variant)}
Similar to EOC-1 with variances in depression angles, the training and test data of four types of SAR target are acquired at $17^{\circ}$ and $30^{\circ}$, respectively.
The data distribution is shown in Table~\ref{data_eoc2}.

\subsubsection{Data descriptions of EOC-3 (version variant)}
The targets of BMP2 and T72 in the MSTAR dataset have different cover different versions (serial numbers). 
As shown in Table~\ref{data_eoc3}, due to the high similarity between different versions of targets within the same class, the differentiation between different types of objects become more challenging.

\subsubsection{Data descriptions of EOC-4 (noise corruption)}
The performance of SAR ATR system is easy to affected by the noise corruption during the perception process.
To simulate real-world scenarios, additive white Gaussian noises with different signal-to-noise ratios (SNRs) are added to SAR images.
The data distribution in this scenario is the same as SOC.

\subsection{General Implementation}
The total number of scattering centers $P$ is set to 40 to make a trade-off between representation power and computational efficiency.
We use an 18-layer ResNet as the backbone to extract deep image features, and a DGCNN with 3 EdgeConv layers to extract scattering features.
The model is trained for 300 epochs with batch size 16 by stochastic gradient descent (SGD) optimizer, and the learning rate is initialized as 0.02.
Before collaboratively conducting the semi-supervised learning of two branches, a 5-epoch warm up period is carried out where the input amplitude images are cropped into $64\times64$.
The probability threshold $\delta$ is set to 0.6 according to the analysis in Subsection \ref{hyper_ana}. 
The other hyperparameters involved in the semi-supervised learning process are kept identical to those in~\cite{li2020dividemix} to ensure a fair comparison.
Specifically, $T$ and $\alpha$ are set to 0.5 and 4, respectively, while the value of $\lambda_u$ is linearly ramped up to 25 over the first 16 epochs.
The number of augmentations $M$ for label refinement and guessing is set to 2, which involves a $64\times64$ random crop within the center $96\times96$ region of the amplitude image, and the random changes in brightness, contrast, and saturation.
The recognition performance is evaluated by the classification accuracy with both symmetric and asymmetric noises across various noise rates.
The generation of symmetric noise is achieved by uniformly changing the true labels for a certain percentage of the training data into other class labels.
This percentage of change in labels refers to the noise rate.
In contrast, for the generation of asymmetric noise, the change of labels for each class is restricted to a specific class based on inter-class similarity (e.g., T62$\leftrightarrow$T72).
The experiments are conducted in the PyTorch framework with Nvidia RTX 3090 GPU support.

\vspace{-0.2cm}
\subsection{Comparison Results}
We compare the performances of our proposed method with state-of-the-art methods of learning with noisy labels:

\textit{SDF-Net}~\cite{liu2022multilevel}: expresses ASC sets at the component level and fuses scattering and deep features in a weighted form.

\textit{Mixup}~\cite{zhang2017mixup}: convexly combines the input data and labels of random sample pairs to eliminate the influence of noisy labels.

\textit{Co-teaching}~\cite{han2018co}: updates each network branch using samples with smaller loss on the other branch.

\textit{Co-teaching+}~\cite{yu2019does}: involves the consideration of model disagreement during the training of each batch.

\textit{JoCoR}\cite{wei2020combating}: introduces a KL divergence-based distance loss to prompt the agreement between two network branches.

\textit{Co-learning}~\cite{tan2021co}: collaboratively accomplishes self-supervised learning and supervised learning on a shared two-head feature encoder.

\textit{DivideMix}~\cite{li2020dividemix}: trains the model in a semi-supervised manner using labeled and unlabeled data divided by GMM.

\textit{ELR}~\cite{liu2020early}: leverages the early-learning phenomenon through a regularization term to enhance the robustness to label noise.

\textit{UNICON}~\cite{karim2022unicon}: rectifies the imbalanced sample selection to enhance the performance with high label noise.

\begin{figure}[]
\centering \vspace{-0.5cm}
\subfloat[Noisy training labels]{\includegraphics[width=0.25\textwidth]{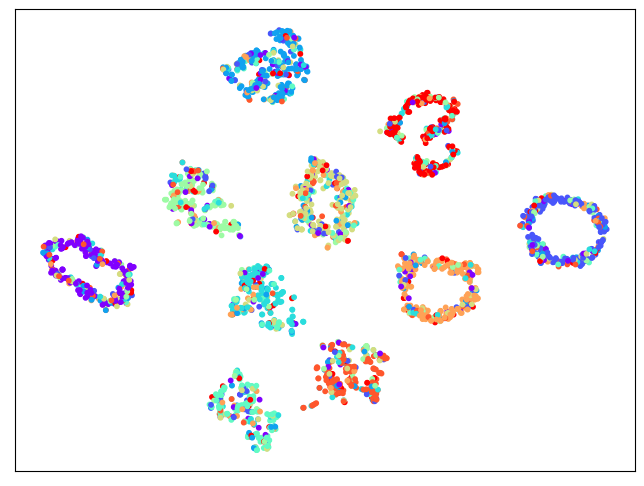}}
\subfloat[True labels]{\includegraphics[width=0.25\textwidth]{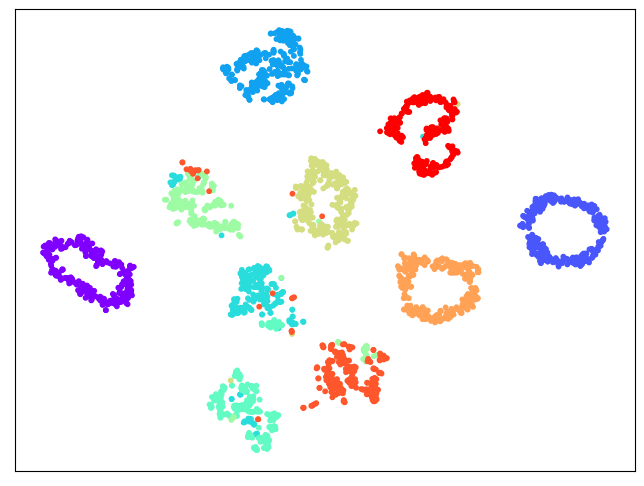}}
\caption{t-SNE visualization of features representations under SOC with $50\% $ symmetric noise. CLSDF can accurately
learn the true data distribution.}
\label{vis_fea} \vspace{-0.35cm}
\end{figure}

\begin{table*}[] \vspace{-0.4cm}
\caption{Comparison with state-of-the-art methods under SOC} 
\label{experiment1} \vspace{-0.1cm}
\centering
\setlength{\tabcolsep}{15pt}
\renewcommand\arraystretch{1.1}
\begin{tabular}{l|cccc|cccc}
\hline \hline
Dataset & \multicolumn{8}{c}{ MSTAR-SOC } \\
\hline \multirow{2}{*}{Methods / Noise rate } & \multicolumn{4}{c|}{ Symmetric } & \multicolumn{4}{c}{ Asymmetric } \\
\cline { 2 - 9} & $20 \%$ & $40 \%$ & $60\%$ & $80 \%$ & $20 \%$ & $30 \%$ & $40 \%$ & $50 \%$ \\
\hline 
Cross-Entropy & 86.37 & 77.88 & 54.88 & 29.08 & 94.54 & 84.75 & 76.92 & 71.50  \\
SDF-Net & 87.09 & 77.28 & 51.78 & 26.37 & 95.07 & 85.72 & 78.39 & 72.68  \\
Mixup & 88.96 & 74.75 & 49.12 & 26.04 & 94.25& 88.29 & 80.29 & 69.71  \\
Co-teaching & 91.72 & 89.36 & 81.49 & 49.94 & 90.60 & 86.34 & 72.10 & 58.27 \\
Co-teaching+ & 92.02 & 87.32 & 82.63 & 49.01 & 90.91 & 86.23 & 77.47 & 59.85 \\
JoCoR & 92.40 & 87.50 & 82.59 & 48.54 & 87.07 & 82.65 & 75.62 & 65.06 \\
Co-learning & 97.26 & 93.09 & 86.51 & 58.23 & 95.77 & 89.56 & 88.03 & 74.99\\
Co-learning (ASC+Image) & 97.63 & 93.67 & 73.25 & 42.78 & 95.07 & 91.89 & 85.60 & 72.60  \\
DivideMix & 97.01& 94.04 & 90.88 & 39.25 & 86.77 & 83.42 & 76.67 & 78.46 \\
DivideMix (ASC+Image) & 98.01 & 94.58 & 92.96 & 63.58 & 92.30 & 91.68 & 84.23 & 82.86  \\
ELR & 98.12& 96.44 & 93.26 & 68.18& 85.58 &82.22 &75.99	 & 59.10 \\
UNICON & \textbf{99.12}& 94.38 & 81.05 & 53.88 & 91.27 & 87.06 & 82.62 & 77.25\\
UNICON (ASC+Image) & 95.11 & 81.79 & 69.75 & 42.88 & 85.67 & 82.38 & 80.71 & 73.33  \\
\hline
CLSDF & \textbf{99.12}&\textbf{98.14} & \textbf{95.53} & \textbf{80.24} & \textbf{95.96} & \textbf{93.58} & \textbf{90.69} & \textbf{85.79} \\ \hline \hline
\end{tabular}
\end{table*}

\begin{figure}[]
\centering \vspace{-0.6cm}
\subfloat[40\% symmetric noise]{\includegraphics[width=0.25\textwidth,height=0.21\textwidth]{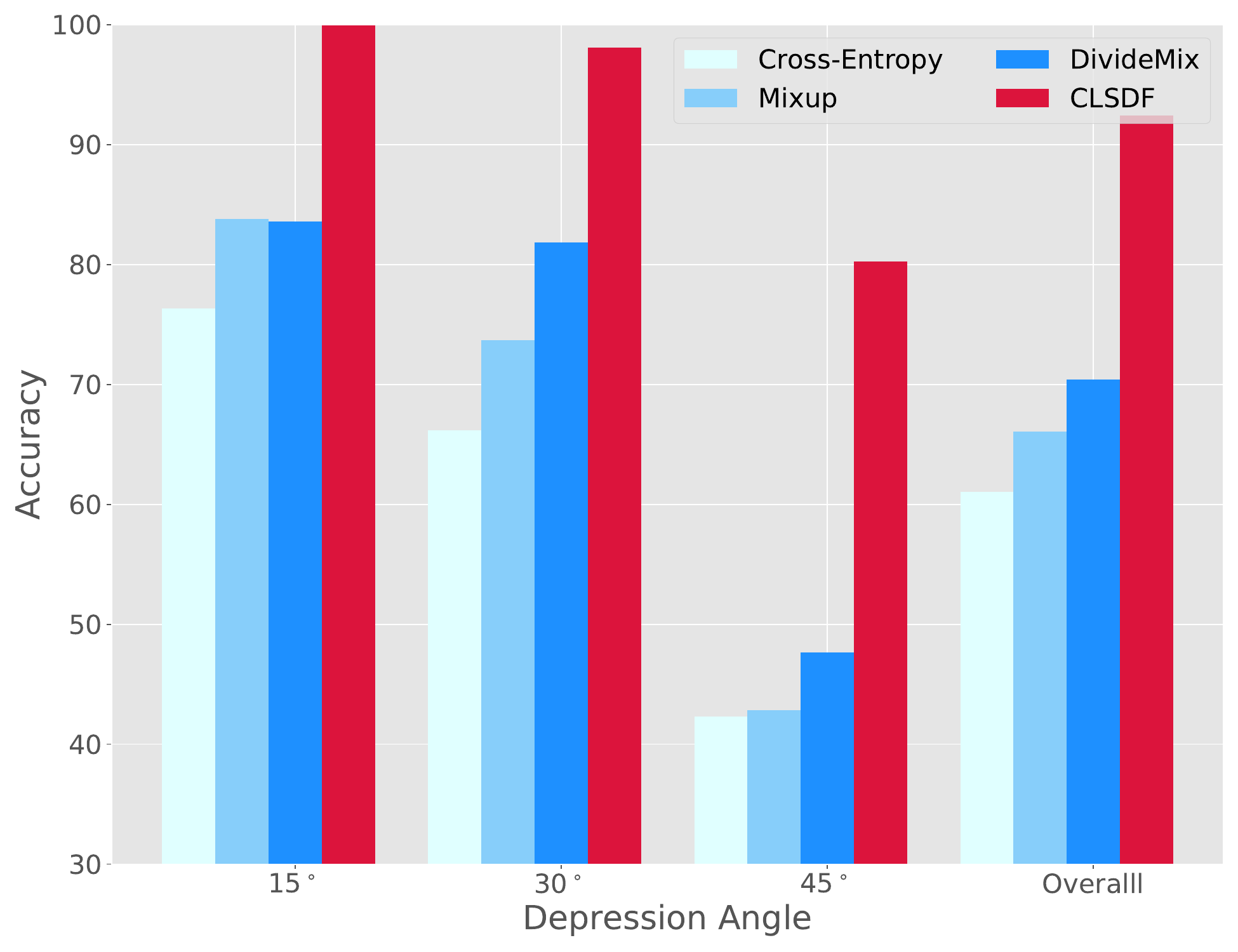}\label{eoc1_40}} 
\subfloat[80\% symmetric noise]{\includegraphics[width=0.25\textwidth,height=0.21\textwidth]{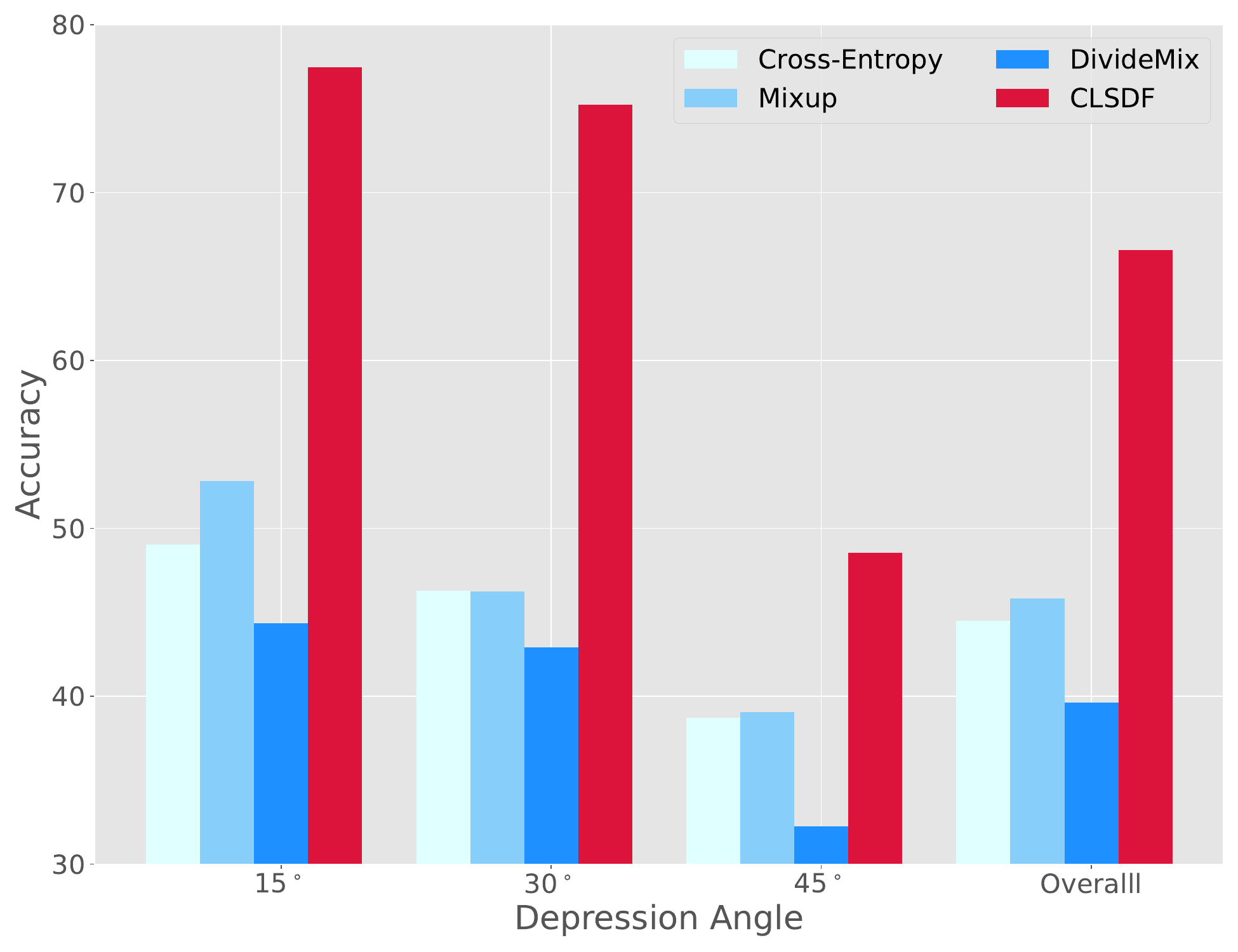}\label{eoc1_80}}
\caption{Comparison with state-of-the-art methods in the test accuracy under EOC-1 in the presence of symmetric noise.}
\label{eoc1} \vspace{-0.3cm}
\end{figure}

\begin{table}[] 
\caption{Comparison with state-of-the-art methods under EOC-2} 
\label{experiment3} \vspace{-0.2cm}
\centering
\setlength{\tabcolsep}{9.9pt}
\renewcommand\arraystretch{1.1}
\begin{tabular}{l|cccc}
\hline \hline
Dataset & \multicolumn{4}{c}{ MSTAR-EOC-2 } \\
\hline \multirow{2}{*}{Methods / Noise rate } & \multicolumn{4}{c}{ Symmetric } \\
\cline { 2 - 5} & $20 \%$ & $40 \%$ & $60\%$ & $80 \%$  \\
\hline 
Cross-Entropy & 91.37 &88.20 & 54.75 &36.36    \\
Mixup & 95.07 &96.12 & 56.07 & 41.02   \\
Co-teaching & 79.13 &76.32 & 79.04 & 62.67   \\
Co-teaching+ & 80.05 & 77.41& 74.42 & 63.28   \\
JoCoR & 83.44 & 79.93 & 63.01 & 42.43   \\
Co-learning & 90.14 & 88.64 & 72.62 & 61.47   \\
DivideMix & 93.13 &72.66 & 71.30 & 42.61   \\
ELR & 83.37 & 80.07 & 76.67 & 63.77   \\
UNICON & 91.34 &79.82 & 67.53 & 48.12   \\
\hline
CLSDF & \textbf{99.74}&\textbf{98.87} & \textbf{92.43} & \textbf{71.57}   \\ \hline \hline
\end{tabular} \vspace{-0.2cm}
\end{table}

\subsubsection{Results under SOC} 
We vary the ratios of symmetric and asymmetric noise from $20 \%-80 \%$ and $20 \%-50 \%$, respectively. 
The results under SOC are shown in Table~\ref{experiment1}, we can see that CLSDF can consistently achieve superior recognition performance than comparison methods.
Compared to other methods, the performance of CLSDF can be maintained stably with the increase of noise ratio.
In addition, the superiority in noise robustness becomes more significant for asymmetric noise, which is more challenging.
The above-mentioned fully demonstrates the effectiveness of our method for SAR ATR in the presence of noisy labels.
Then, we find that the recognition performance of the sample selection-based methods generally outperform the regularization-based methods.
This proves the potential applicability of this type of learning framework for SAR ATR with noisy labels.
However, their robustness to changes in noisy labels with type and ratio falls short of expectations.
For in-depth exploration of the applicability of existing methods, we also evaluate their performance with the fused scattering and deep features (marked as ASC+Image).
The recognition performance of DivideMix can gain a certain improvement based on the enriched representation.
Nevertheless, the performance of other methods remains unsatisfactory and even suffers from significant degradation.
Based on the above analysis, the cause of the inapplicability of existing methods for SAR ATR with noisy labels is twofold.
First, the single visual characteristic fails to provide comprehensive descriptions of the target. 
Second, the adequacy of learning from the enriched representation needs to be further improved.

Besides the recognition accuracy, we visualize the learned feature representations of CLSDF by t-SNE~\cite{vanvisualizing} with $50\% $ symmetric noise. 
The adopted feature representations correspond to the integrated deep and scattering features of the training data, which are extracted by the multi-model feature fusion framework.
As shown in Fig.~\ref{vis_fea}, the representations form ten distinct clusters that correspond to the true class labels, which demonstrates that CLSDF can robustly learn the underlying distribution in the presence of noisy labels.

\begin{table*}[]  \vspace{-0.6cm}
\caption{Comparison with state-of-the-art methods under EOC-3} 
\label{experiment4} 
\centering
\setlength{\tabcolsep}{5pt}
\renewcommand\arraystretch{1.1}
\begin{tabular}{|c|c|c|c|c|c|c|c|c|c|c|c|c|c|}
\hline
\multicolumn{2}{|c|}{Dataset}&\multicolumn{12}{c|}{ MSTAR-EOC-3 ($30\%$ Symmetric) } \\
\hline \multirow{2}{*}{ Class } & \multirow{2}{*}{$\begin{array}{c}\text {Serial} \\
\text {Number}\end{array}$} & \multicolumn{4}{c|}{Cross-Entropy} & \multicolumn{4}{c|}{DivideMix}& \multicolumn{4}{c|}{ CLSDF }\\
\cline { 3- 14} & & BMP2 & $\mathrm{BRDM} 2$ & $\text { BTR70 }$ & $\mathrm{T} 72$& BMP2 & $\mathrm{BRDM} 2$ & $\text { BTR70 }$ & $\mathrm{T} 72$& BMP2 & $\mathrm{BRDM} 2$ & $\text { BTR70 }$ & $\mathrm{T} 72$  \\
\hline \multirow{2}{*}{ BMP2 } &9566 &324&27&20&57 &315&43&23&47 &347&6&30&45  \\
\cline { 2- 14} & C21 &210&48&48&123 &206&960&31&102 &282&18&49&80\\
\hline \multirow{5}{*}{ T72 } &812 &117&33&28&248 &91&49&33&253 &76&2&28&320 \\
\cline { 2- 14} & A04 &159&73&50&291 &95&137&27&314 &76&23&11&463\\
\cline { 2- 14} & A05 &116&86&31&340 &62&136&28&347 &39&26&9&499\\
\cline { 2- 14} & A07 &159&68&26&320 &91&131&30&321 &30&29&4&510 \\
\cline { 2- 14} & A10 &121&72&29&345 &54&127&30&356 &44&26&5&492 \\
\hline \multicolumn{2}{|c|}{ Average } & \multicolumn{4}{c|}{58.22}&  \multicolumn{4}{c|}{59.18} &  \multicolumn{4}{c|}{\textbf{81.62}}\\
\hline
\end{tabular}
\end{table*}

\begin{figure*}[]
\centering  \vspace{-0.5cm}
\subfloat[20\% symmetric noise]{\includegraphics[width=0.32\textwidth]{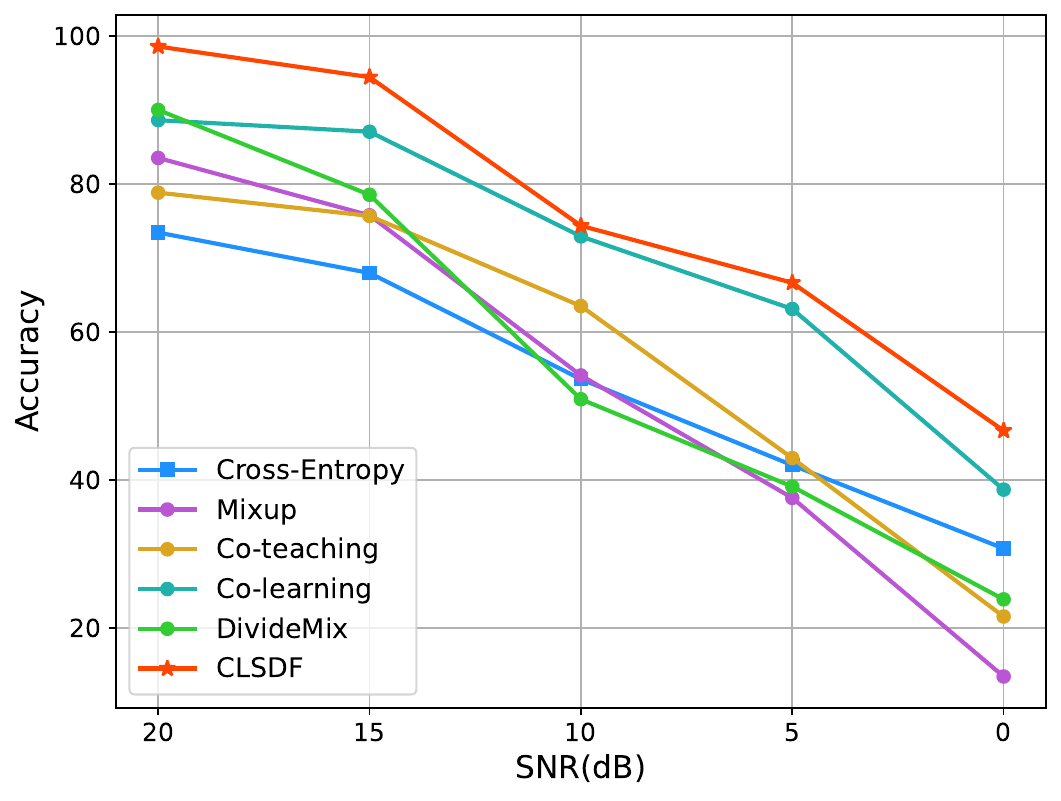}\label{eoc4_1}}\hspace{0.2cm}
\subfloat[40\% symmetric noise]{\includegraphics[width=0.32\textwidth]{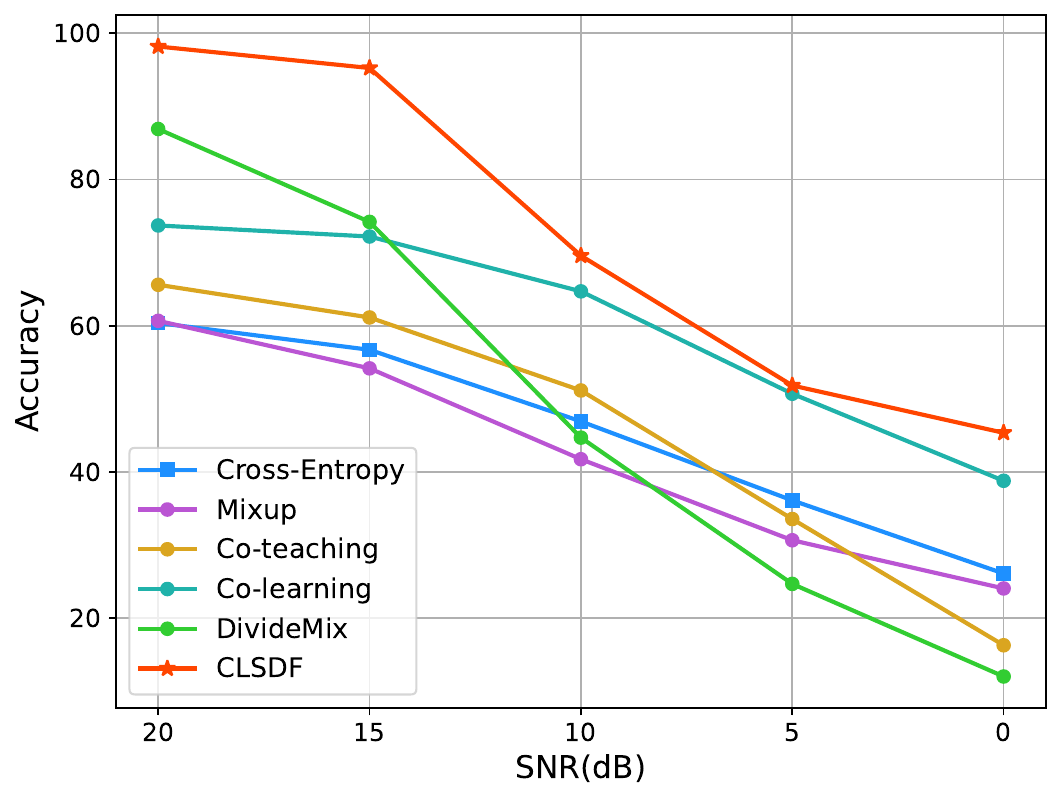}\label{eoc4_2}}\hspace{0.2cm}
\subfloat[40\% asymmetric noise]{\includegraphics[width=0.32\textwidth]{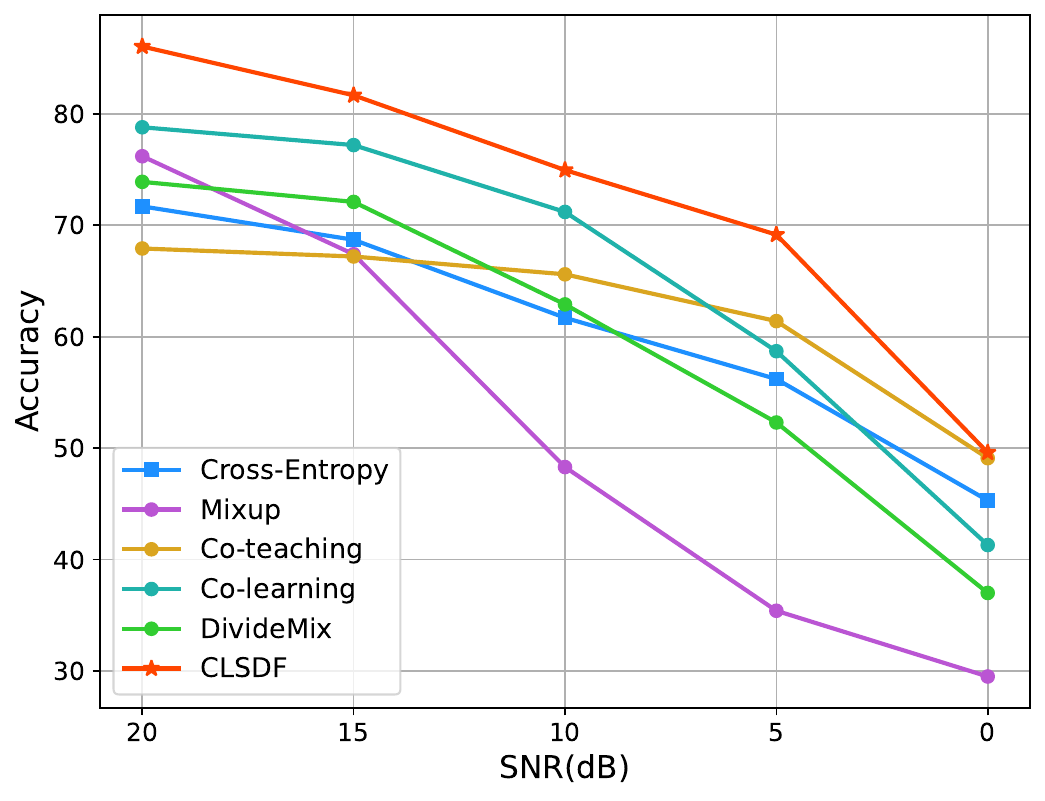}\label{eoc4_3}}
\caption{Performances of different methods under EOC-4 with different noisy labels.}
\label{eoc4} \vspace{-0.2cm}
\end{figure*}

\subsubsection{Results under EOC-1 and EOC-2} 
The experiment results with variances in the depression angle are reported in Fig.~\ref{eoc1} and Table~\ref{experiment3}, corresponding to EOC-1 and EOC-2, respectively.
First, the results in both tables show that our method can still achieve outstanding performance in the presence of noisy labels when facing differences in observation angles.
Then, according to the results shown in Fig.~\ref{eoc1} under EOC-1, we find that the recognition performances tend to decline with increasing differences in depression angle.
Furthermore, this decay is more significant in the presence of a larger rate of noisy labels.
Besides, as demonstrated in Table~\ref{experiment3} under EOC-2 with only four types of target, the recognition performance of each method can achieve better performance at lower noise rates than SOC. 
However, the recognition performances at higher noise rates become lower than SOC.
This indicates that the inclusion of variances in the depression angle will bring further challenges to SAR ATR with noisy labels.

\subsubsection{Results under EOC-3} 
We compare the proposed method with cross-entropy and DivideMix under EOC-3 with $30\%$ symmetric noise, which can be regarded as CNN baseline and typical sample selection-based method, respectively.
The performance with version variant is shown in detail through the corresponding confusion matrices in Table~\ref{experiment4}.
First, conventional CNN architecture causes significant misclassification of different versions of targets.
Then, DivideMix similarly struggles with accurately recognizing different versions of targets, which is mainly due to insufficient feature representation and imprecise data division.
In contrast, the proposed method can effectively avoid the inter-class confusion with version variant in the presence of noisy labels.

\subsubsection{Results under EOC-4} 
To validate the effectiveness of the proposed method
with noise corruptions from both data annotation and data acquisition, we evaluate the performance under EOC-4 with different types and ratios of noisy labels.
Specifically, the training data for this condition is the same as SOC, and different levels of noise are imposed on the magnitude image of the test data under SOC, making the SNR varying between $[20,15,10,5,0]$ dB.
According to the results shown in Fig.~\ref{eoc4}, we can see that CLSDF outperforms other method across various types and levels of noise.
In addition, the performance of each method decays with increasing levels of noise interference, mainly due to the fact that the presence of noisy labels can severely affect the discriminability of classification decision boundaries.

\subsubsection{Results on other target types}
\begin{figure*}[]
\centering  \vspace{-0.5cm}
\subfloat[]{\includegraphics[width=0.25\textwidth]{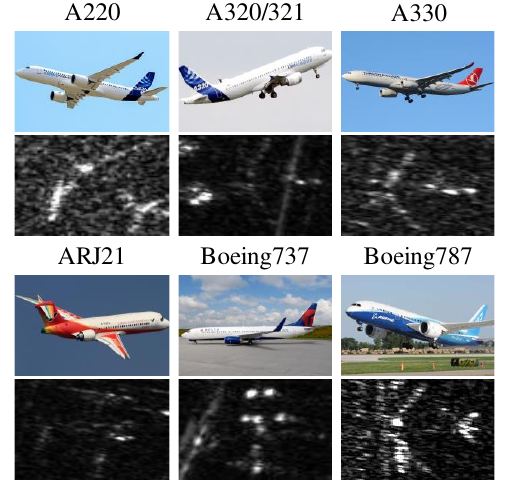}\label{air_data}}
\subfloat[Co-teaching]{\includegraphics[width=0.25\textwidth]{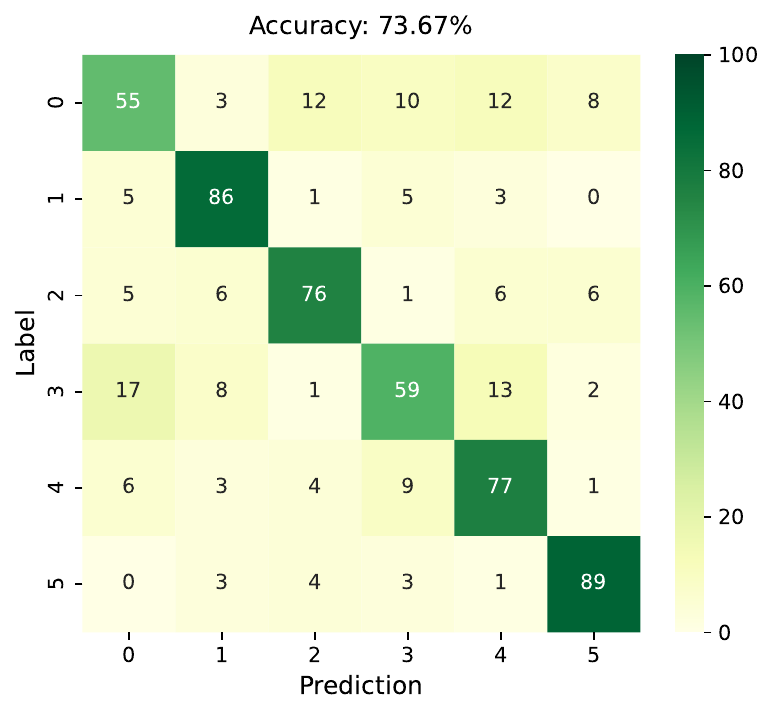}\label{air_1}}
\subfloat[DivideMix]{\includegraphics[width=0.25\textwidth]{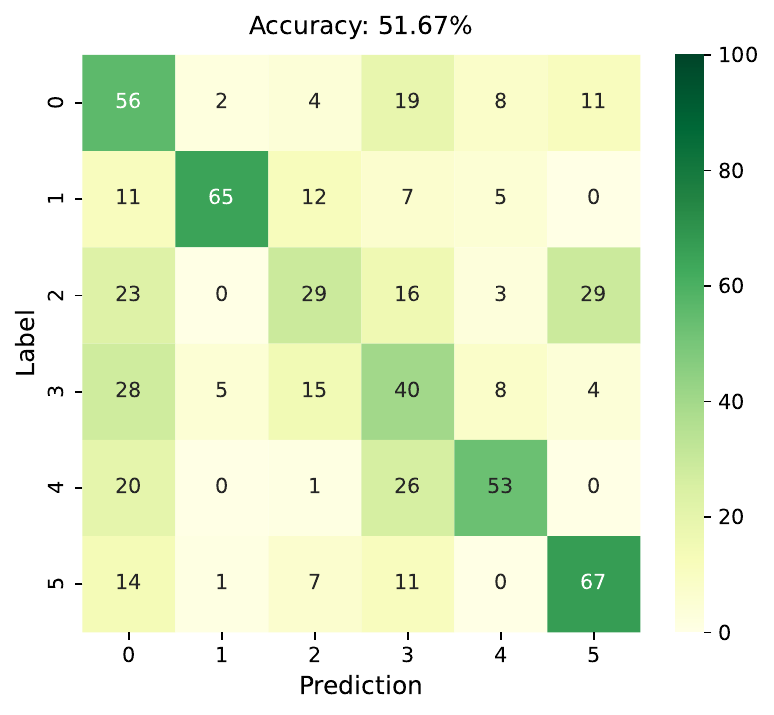}\label{air_2}}
\subfloat[CLSDF]{\includegraphics[width=0.25\textwidth]{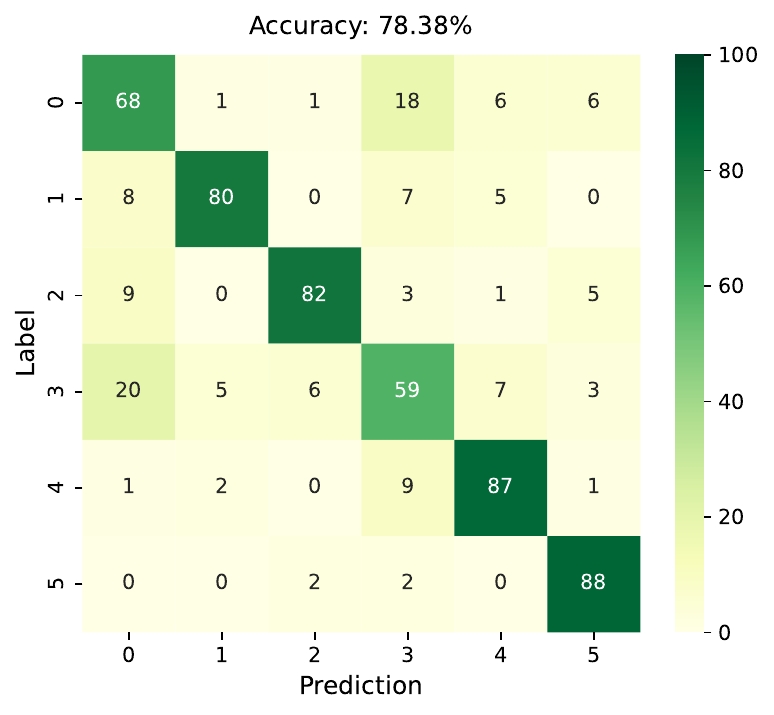}\label{air_3}}
\caption{(a) Optical and SAR images of each class in the SAR-ACD dataset, and the confusion matrices of (b) Co-teaching, (c) DivideMix, and (d) CLSDF.}
\label{sar_air} \vspace{-0.2cm}
\end{figure*}

Beyond the validation on the MSTAR dataset for vehicle recognition, we further investigate the effectiveness of the proposed collaborative learning framework in noise-robust training for other target types.
Specifically, we conduct experiments on the SAR-ACD~\cite{sunscan} dataset, a real-world SAR aircraft recognition dataset collected by the GF-3 satellite.
Compared to MSTAR, SAR-ACD presents greater challenges, as the aircraft targets are acquired across different airports, exhibiting more complex structural variations and large size differences.
For the released six categories of civil aircraft visualized in Fig.~\ref{air_data}, each contains about 500 samples, with 100 samples per category used for testing.
The scattering features in SAR-ACD are extracted from strong scattering points following their original implementation.
We compare the performance of different methods with $50\%$ symmetric noise, and the recognition results are reported by confusion matrices, where the corresponding classification accuracy is displayed above each.
As shown in Figs.~\ref{air_1}-\ref{air_3}, CLSDF can still achieve advanced performance, demonstrating the general applicability of the proposed method across different SAR ATR scenarios.

\begin{table}[] \vspace{-0.2cm}
\caption{Ablation experiments to prove the effectiveness of multi-model feature extraction, class-wise sample selection, and joint distribution alignment}
\label{table_ablation1} \vspace{-0.2cm}
\center
\renewcommand\arraystretch{1.15}
\begin{tabular}{l|ccc|cc}
\hline \hline Methods &MMFE&CWSS&JDA & Sym. & Asym. \\  \hline
DivideMix &   &   &  & 94.04 & 76.67 \\
w/o CWSS and JDA & \checkmark &   &  & 94.58 & 84.23 \\
w/o JDA & \checkmark & \checkmark &   & 95.12 & 88.29 \\
w/o CWSS & \checkmark &   & \checkmark& 95.90 & 88.66 \\
w/o MMFE &  &  \checkmark & \checkmark& 96.40 & 86.51 \\ \hline
CLSDF &  \checkmark & \checkmark & \checkmark & \textbf{98.14} &  \textbf{90.69} \\
\hline \hline
\end{tabular} \vspace{-0.2cm}
\end{table}

\subsection{Ablation Studies}\label{ex_ab}
\subsubsection{Contribution of each main part}
The proposed method belongs to the type of sample selection-based method of learning with noisy labels.
In terms of model architecture design, CLSDF can be viewed as an extended version of DivideMix, a advanced image-based collaborative learning framework containing two network branches.
The involved improvements refer to three main parts, i.e., the multi-model feature extraction~(MMFE), the class-wise sample selection~(CWSS), and the joint distribution alignment~(JDA).
We conduct ablation experiments with both symmetric and asymmetric noise at $40\%$ noise rate on SOC to prove the contribution of each module to noise-robust learning.

The performance are reported in Table~\ref{table_ablation1}.
First, we find that only providing richer representations during the learning process through MMFE can lead to a certain degree of performance improvement.
This emphasizes the necessity of integrating scattering and deep features for SAR ATR with noisy labels. 
However, there remains a large room for performance improvement, especially when dealing with symmetric noise.
Then, the individual utilization of both CWSS and JDA can further improve the recognition performance based on the fusion feature extracted by MMFE.
Moreover, the performance of our method based only on amplitude image also outperforms the baseline framework, demonstrating the generalizability of CWSS and JDA to image-based noise robust learning.
Finally, combining all these parts into the proposed CLSDF can achieve the best recognition performance.

\begin{figure}[]
\centering \vspace{-0.3cm}
\subfloat[Symmetric noise]{\includegraphics[width=0.25\textwidth,height=0.21\textwidth]{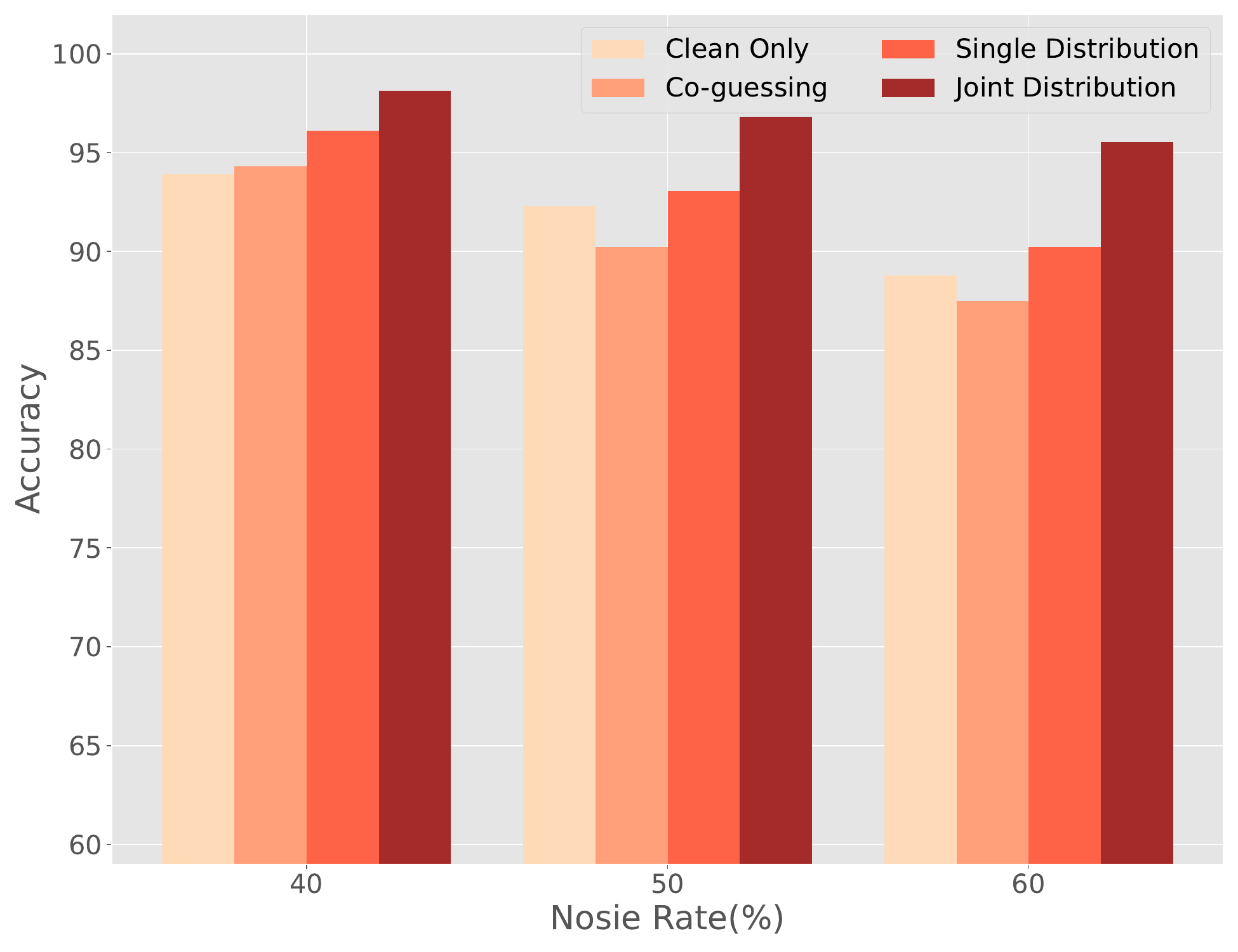}\label{ab2_1}} 
\subfloat[Asymmetric noise]{\includegraphics[width=0.25\textwidth,height=0.21\textwidth]{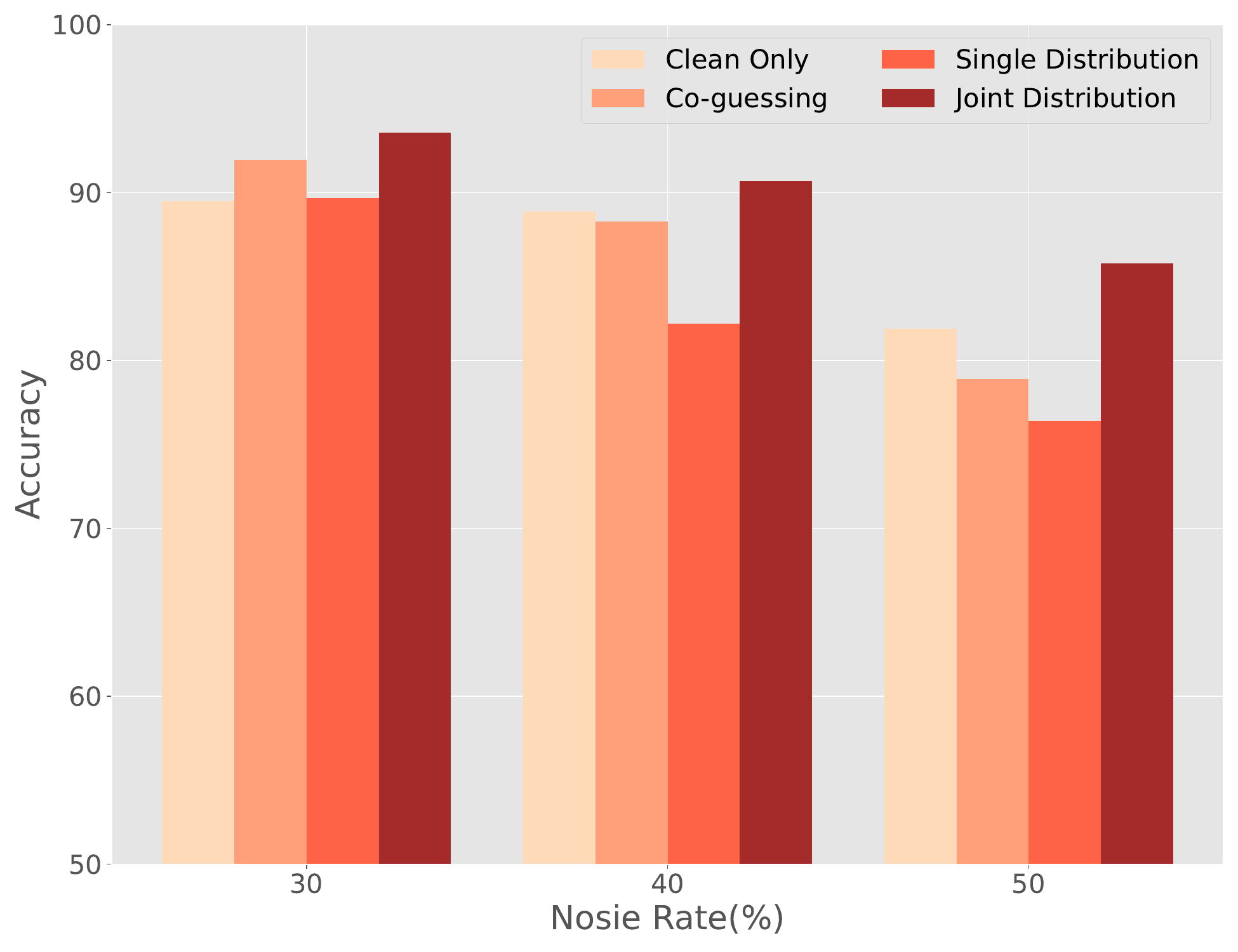}\label{ab2_2}}
\caption{Ablation experiments for the joint distribution alignment in the presence of different types and rates of noisy labels.}
\label{ablation2} \vspace{-0.2cm}
\end{figure}

\subsubsection{Effectiveness of joint distribution alignment}
During the semi-supervised learning process of each branch, the utilization of data in the noisy subset is heavily dependent on the reliability of label guessing results.
However, the labeling guesses results are biased towards over-accumulation on certain classes, which deviates from the underlying category information.
We propose a joint distribution alignment strategy to alleviate this problem by encouraging the joint distribution of clean and guessed labels to match the marginal class distribution.
To prove the effectiveness of the joint distribution alignment strategy, we compare the performance of the proposed method on SOC with different strategies for semi-supervised learning.

The corresponding results on SOC with various ratios of symmetric and asymmetric noise are shown in Figs.~\ref{ab2_1} and~\ref{ab2_2}.
The Clean Only represents abandoning the noisy subset and using only the clean subset for supervised learning.
The Co-guessing refers to generating guessed labels without distribution alignment.
The Single Alignment refers to enforcing the guessed labels of labeled data to match the distribution of labeled data~\cite{berthelot2019remixmatch}.
According to the results, semi-supervised learning on guessed labels without distribution alignment achieves even worse performance than using only labeled data in the clean subset at high noise ratios.
Then, the Single Alignment can improve the performance in the presence of symmetric noise, but fails to handle asymmetric noise, mainly due to the non-uniform distribution of class labels.
In contrast, the proposed joint distribution alignment strategy can stably improve the recognition performance across various types and ratios of noisy labels.

\subsection{Parameter Analysis}\label{hyper_ana}
\subsubsection{Data division}
Among the hyperparameters involved in the proposed method, the probability threshold $\delta$ plays a crucial role in data division, which affects the subsequent semi-supervised learning process.
We conduct experiments to analyze the impact of different values of $\delta$ on the precision of data division.
Specifically, we vary $\delta$ from 0.5 to 0.8 during the sample selection process with $50\%$ symmetric noise.
The effectiveness of data division is measured by the division accuracy of correctly labeled samples, and the division error of mislabeled samples of overall two branches.

\begin{figure}[]
\centering \vspace{-0.3cm}
\subfloat[Division accuracy]{\includegraphics[width=0.25\textwidth,height=0.17\textwidth]{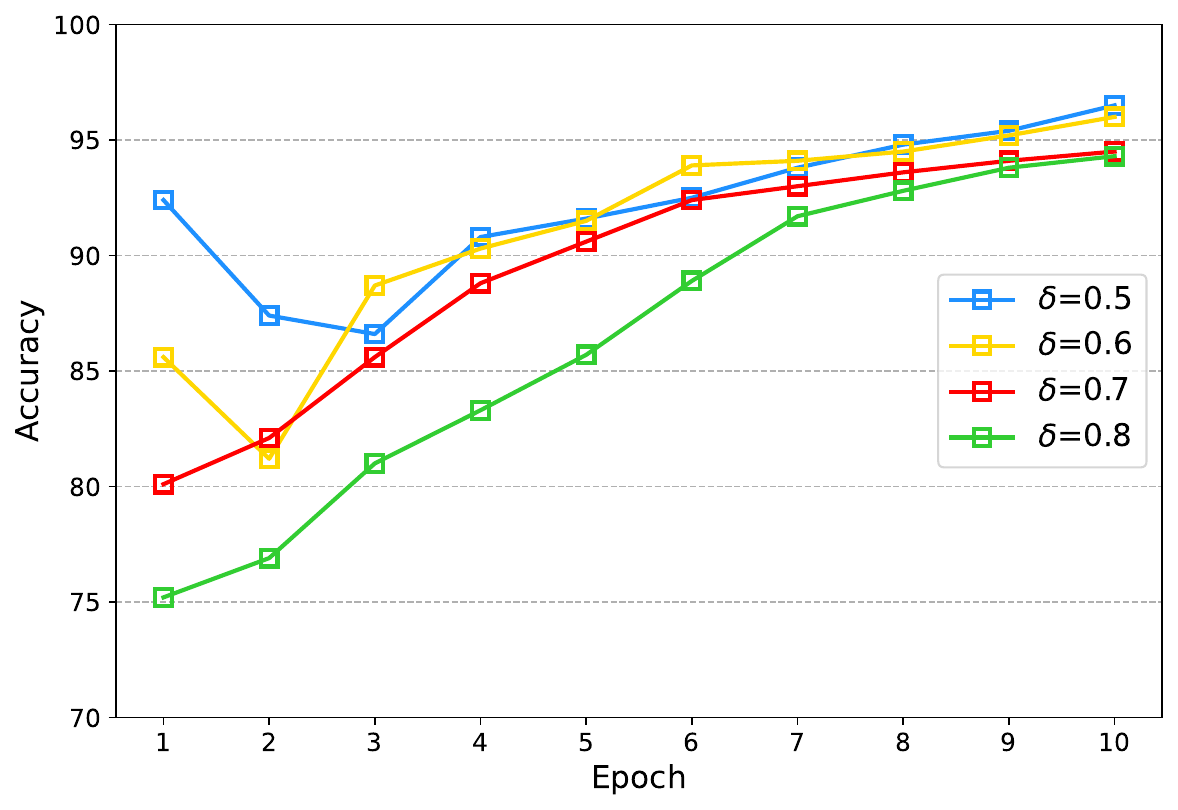}\label{hyper1}}
\subfloat[Division error]{\includegraphics[width=0.25\textwidth,height=0.17\textwidth]{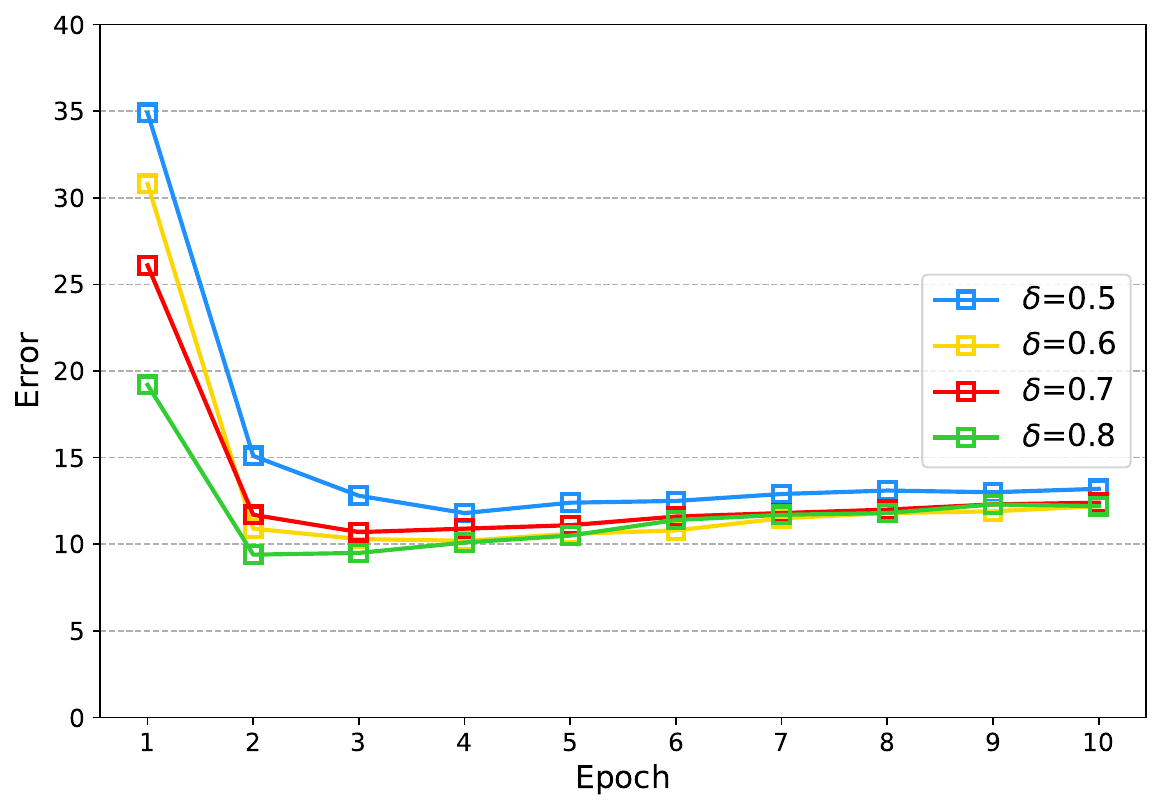}\label{hyper2}}
\caption{The measurement of data division for different settings of $\delta$.}
\label{hyper} \vspace{-0.2cm}
\end{figure}

The measurements of data division quality during the first 10 epochs of the semi-supervised learning process after warm up are reported in Fig.~\ref{hyper}.
In general, the data division precision of the proposed method tends to remain stable across a wide range of $\delta$.
Besides, according to the results shown in Fig.~\ref{hyper1}, we can see that a smaller value of $\delta$ leads to a lenient condition for clean sample selection, thus more correctly labeled samples will be divided into the clean subset.
In contrast, as shown in Fig.~\ref{hyper2}, a higher value of $\delta$ can reduce the number of mislabeled samples to be wrongly divided into the clean subset.
Therefore, setting $\delta$ to a moderate value of 0.6 is more reasonable, as it can balance the division accuracy of correctly labeled samples and the division error of mislabeled samples.

\begin{figure}[]
\centering \vspace{-0.3cm}
\subfloat[] {\includegraphics[width=0.25\textwidth,height=0.17\textwidth]{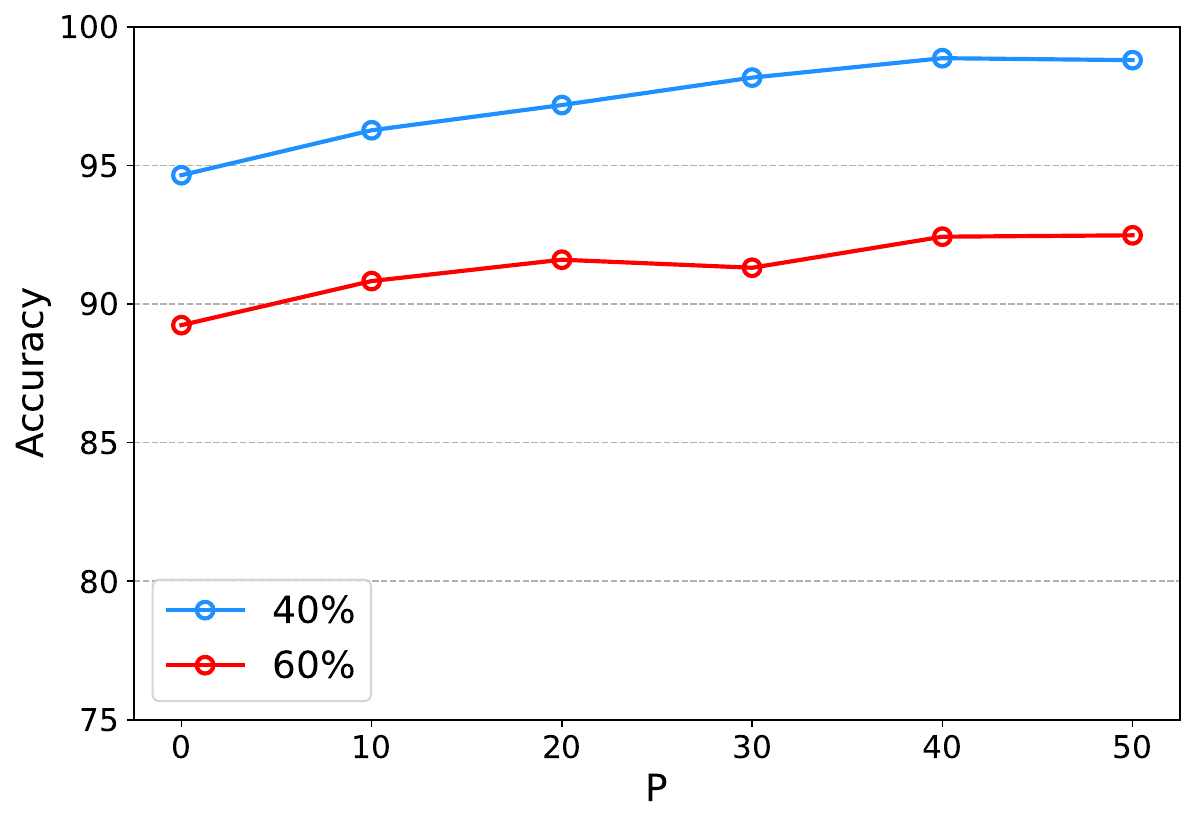}\label{struc1}}
\subfloat[] {\includegraphics[width=0.25\textwidth,height=0.17\textwidth]{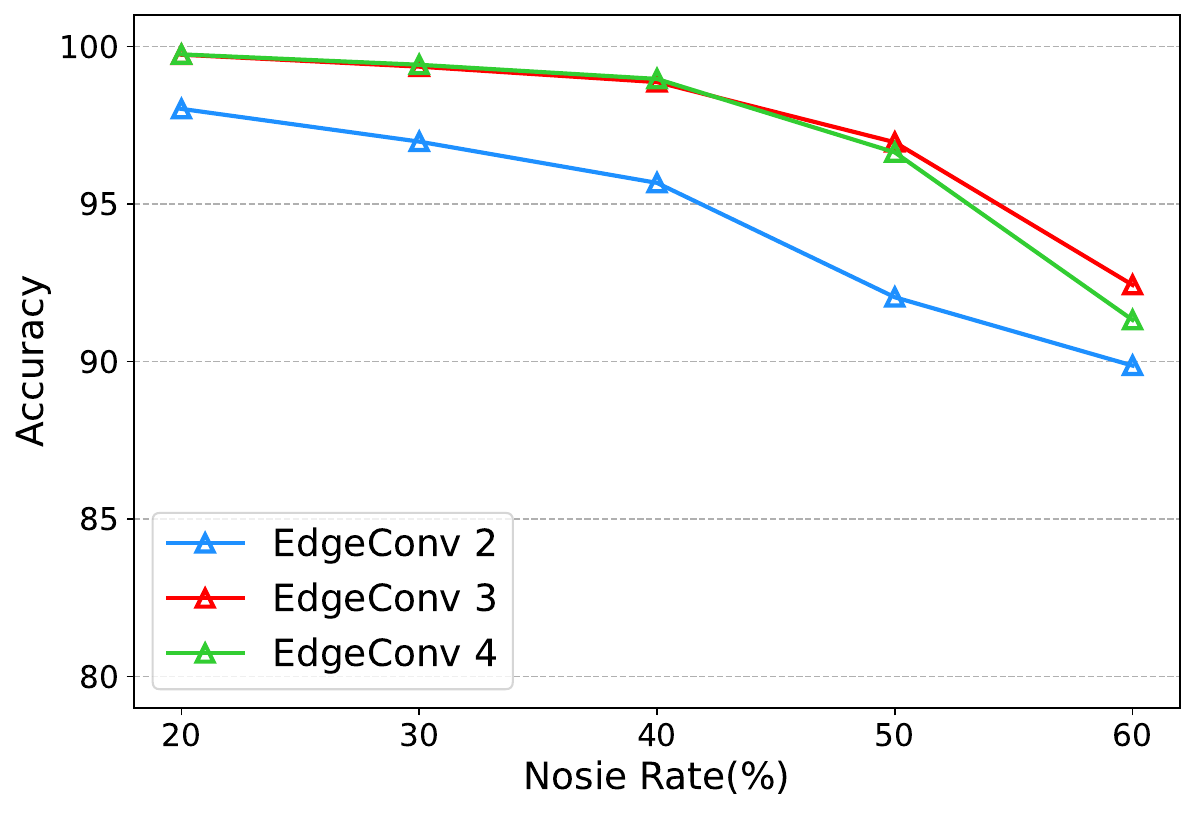}\label{struc2}}
\caption{Analysis of parameter settings regarding model and data structures.}
\label{structure} \vspace{-0.2cm}
\end{figure}

\subsubsection{Model and data structure}
There also exist parameters about model and data structures, i.e., the number of EdgeConv layers and scattering centers $P$.
We investigate the influence of these two parameters on the recognition performance under EOC-2 with different ratios of symmetric noise. 

First, we set the EdgeConv layers to 3 and vary $P$ from 10 to 50.
As shown in Fig.~\ref{struc1}, the recognition performance can be improved by increasing $P$ in a certain range as the physical characteristics are better reflected.
However, the performance improvement tends to saturate regardless of the increase in the number of scattering centers.
Therefore, setting the number of scattering centers to 40 can achieve a better balance between accuracy and computational costs. 
Then, we observe in Fig.~\ref{struc2} that the recognition performance decreases significantly when the number of EdgeConv layers is set to 2, which indicates insufficient capture of topological relations. 
In contrast, increasing the number of EdgeConv layers to 4 does not lead to significant performance improvements, but even increases the risk of overfitting when the noise ratio is high. 
Overall, we set the number of scattering centers and EdgeConv layers to 40 and 3, respectively.

\section{Conclusion}\label{sec5}
The robustness to noisy labels is critical to guarantee the reliability of SAR ATR systems.
In this paper, we have proposed CLSDF to achieve noise-robust SAR ATR.
The proposed CLSDF integrates scattering and deep features from diverse perspectives, thereby overcoming the limitations of existing image-based methods when applied to complex SAR data.
Multiple class-wise GMMs are employed to improve the precision of dividing data into clean and noisy subsets.
Moreover, the over-accumulation of bias in label guessing results is alleviated by the joint distribution alignment strategy for clean and guessed labels.
Thorough experiments across various types and ratios of label noise under different operating conditions have been conducted to validate the effectiveness of the proposed CLSDF.
Compared with existing state-of-the-art methods, CLSDF can achieve superior recognition performance.
In the future, we intend to explore learning with noisy labels in more complex scenarios, such as few-shot or open-set settings, which align more closely with the deployment requirements of real-world SAR ATR systems.

\bibliographystyle{IEEEtran}
\bibliography{CLSDF}

\end{document}